\documentclass[sigconf,nonacm]{acmart}

\usepackage{subfigure}
\definecolor{pykalegreen}{RGB}{63, 127, 84}
\newcommand{\codetxt}[1]{\texttt{#1}}
\newcommand{\kalegreen}[1]{\textcolor{pykalegreen}{\textbf{#1}}}
\usepackage[frozencache=true,cachedir=.,newfloat]{minted}
\usepackage{newfloat}
\DeclareFloatingEnvironment[
    fileext=los,
    listname={List of Codes},
    name=Scheme,
    placement=tbhp,
    within=section,
]{code}
\SetupFloatingEnvironment{listing}{name=Code}

\AtBeginDocument{%
  \providecommand\BibTeX{{%
    \normalfont B\kern-0.5em{\scshape i\kern-0.25em b}\kern-0.8em\TeX}}}

\setcopyright{acmcopyright}
\copyrightyear{2021}
\acmYear{2021}
\acmDOI{10.1145/1122445.1122456}

\begin{document}

\title{PyKale: Knowledge-Aware Machine Learning from Multiple Sources in Python}

\author{Haiping Lu}
\affiliation{%
  \institution{The University of Sheffield}
  \city{Sheffield}
  \country{United Kingdom}
}
\email{h.lu@sheffield.ac.uk}

\author{Xianyuan Liu}
\affiliation{%
  \institution{The University of Sheffield}
  \city{Sheffield}
  \country{United Kingdom}}
\email{xianyuan.liu@sheffield.ac.uk}

\author{Robert Turner}
\affiliation{%
  \institution{The University of Sheffield}
  \city{Sheffield}
  \country{United Kingdom}}
\email{r.d.turner@sheffield.ac.uk}

\author{Peizhen Bai}
\affiliation{%
  \institution{The University of Sheffield}
  \city{Sheffield}
  \country{United Kingdom}}
\email{PBai2@sheffield.ac.uk}

\author{Raivo E Koot}
\affiliation{%
  \institution{The University of Sheffield}
  \city{Sheffield}
  \country{United Kingdom}}
\email{rekoot1@sheffield.ac.uk}

\author{Shuo Zhou}
\affiliation{%
  \institution{The University of Sheffield}
  \city{Sheffield}
  \country{United Kingdom}}
\email{SZhou20@sheffield.ac.uk}

\author{Mustafa Chasmai}
\affiliation{%
 \institution{Indian Institute of Technology, Delhi
}
\city{New Delhi}
 \country{India}}
\email{cs1190341@iitd.ac.in}

\author{Lawrence Schobs}
\affiliation{%
  \institution{The University of Sheffield}
  \city{Sheffield}
  \country{United Kingdom}}
\email{laschobs1@sheffield.ac.uk}

\renewcommand{\shortauthors}{Lu, et al.}

\begin{abstract}
Machine learning is a general-purpose technology holding promises for many interdisciplinary research problems. However, significant barriers exist in crossing disciplinary boundaries when most machine learning tools are developed in different areas separately. We present \textit{Pykale} -- a Python library for knowledge-aware machine learning on graphs, images, texts, and videos to enable and accelerate interdisciplinary research. We formulate new green machine learning guidelines based on standard software engineering practices and propose a novel \textit{pipeline}-based application programming interface (API). PyKale focuses on leveraging knowledge from multiple sources for accurate and interpretable prediction, thus supporting multimodal learning and transfer learning (particularly domain adaptation) with latest deep learning and dimensionality reduction models. We build PyKale on PyTorch and leverage the rich PyTorch ecosystem. Our pipeline-based API design enforces standardization and minimalism, embracing \textit{green machine learning} concepts via reducing repetitions and redundancy, reusing existing resources, and recycling learning models across areas. We demonstrate its interdisciplinary nature via examples in bioinformatics, knowledge graph, image/video recognition, and medical imaging. 
\end{abstract}


\maketitle

\section{Introduction}%

Machine learning is the cornerstone technology for artificial intelligence (AI), driving many advances in our everyday lives and industrial sectors. AI research becomes more and more interdisciplinary as many problems rely on expertise from various domains. We have also witnessed many machine learning models transverse different research areas and disciplines. For example, the success of convolutional neural networks (CNNs) \cite{Krizhevsky2012ImageNetCW} has spread from computer vision to graph analysis via graph convolutional networks (GCN) \cite{Kipf:2016tc} and medical imaging via U-net \cite{ronneberger2015unet}, and transformers \cite{vaswani2017attention} developed in natural language processing (NLP) have become a hot topic in solving vision tasks \cite{wang2018non, carion2020end, dosovitskiy2021an}.

With rapid development and growing interests in machine learning, many researchers hope to solve real-world interdisciplinary problems using machine learning. However, even with the popularity of open-source software and high-level scripting language such as Python, navigating the abundant choices and variety of machine learning software is not trivial. Researchers often run into barriers when adapting a machine learning tool for a new task of their interest. Solving complex real-world problems in practice often involve analyzing multiple sources of data, e.g., multiple modalities, multiple domains, and multiple knowledge bases. Most machine learning software packages are developed with a specific domain of application in mind. While popular generic packages such as PyTorch and TensorFlow support multiple domains and are not tailored for a specific domain, their focus on generic frameworks makes them inadequate to directly support interdisciplinary research where both flexible configurations and high-level integration are important.

\begin{figure*}[!t]
  \centering
  \subfigure[Pipeline-based API]{	\label{fig_kaleapi}
		\includegraphics[trim={0 4mm 0 4mm},clip,width=0.8\linewidth]{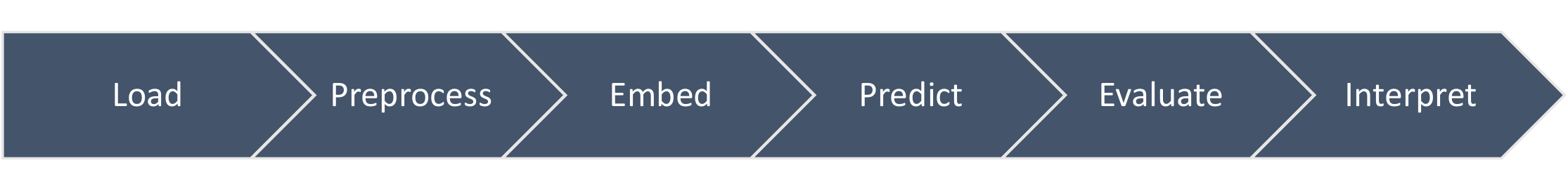}
  }
  \subfigure[Digit classification pipeline]{	\label{fig_digitkaleapi}
		\includegraphics[trim={0 4mm 0 4mm},clip,width=0.8\linewidth]{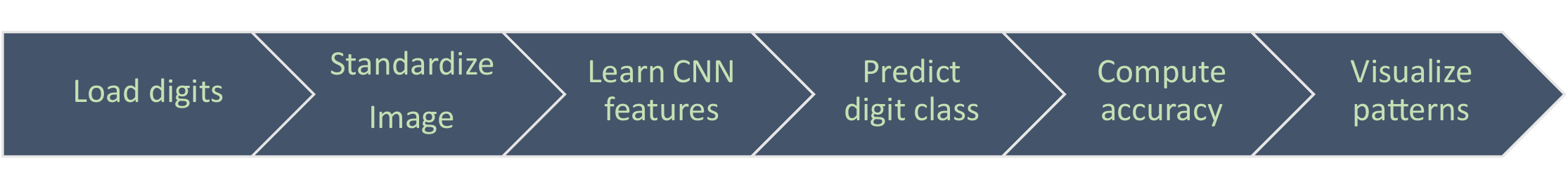}
  }  
  \subfigure[Drug target interaction prediction pipeline]{	\label{fig_drugkaleapi}
		\includegraphics[trim={0 4mm 0 4mm},clip,width=0.8\linewidth]{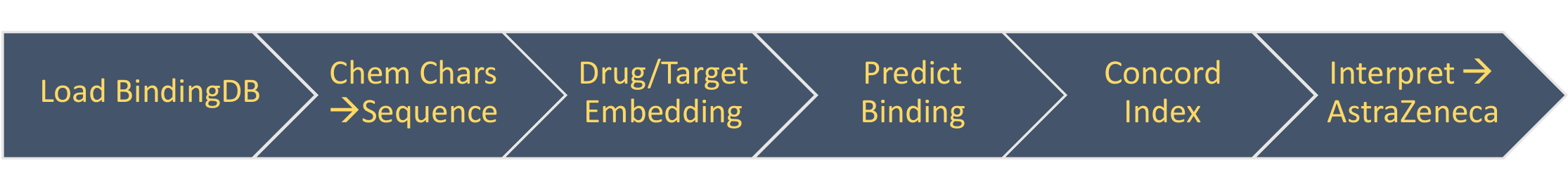}
  }  
  \caption{The proposed pipeline-based API in PyKale and two real-world examples.}\label{pykale_pipeline}
\end{figure*}

\begin{figure}[!t]
  \centering
  \subfigure[Reduce]{\label{fig_reduce}
		\includegraphics[width=0.12\linewidth]{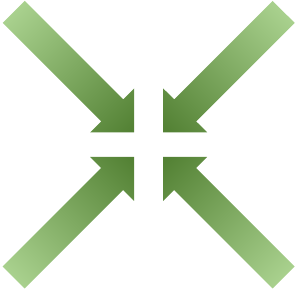}
  }  
  \subfigure[Reuse]{\label{fig_reuse}
		\includegraphics[width=0.17\linewidth]{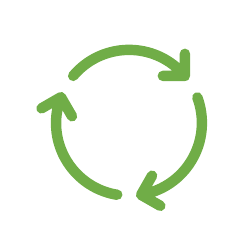}
  }  
  \subfigure[Recycle]{\label{fig_recycle}
		\includegraphics[width=0.15\linewidth]{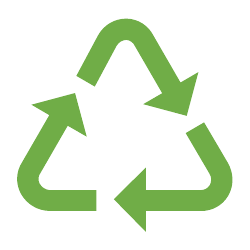}
  }    
  \subfigure[The PyKale logo]{\label{pykale_logo}
		\includegraphics[width=0.45\linewidth]{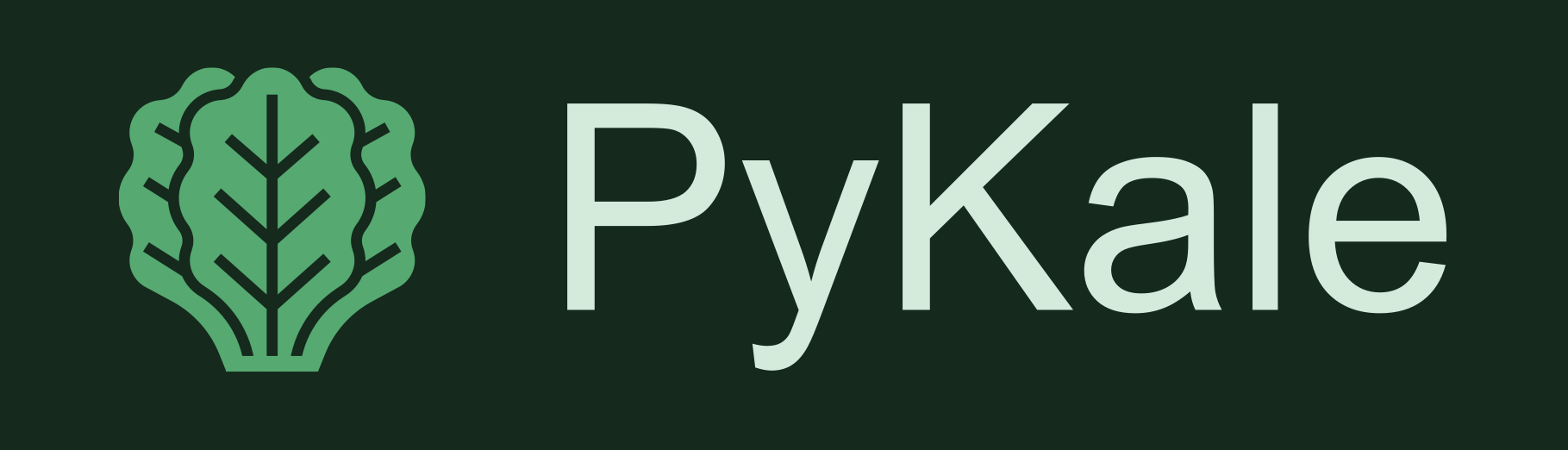}
  }
  \caption{\kalegreen{Green machine learning} concepts in PyKale.}\label{pykale_green}
\end{figure}

In this paper, we propose \textit{PyKale}, an open-source Python library to enable and accelerate interdisciplinary research via knowledge-aware multimodal learning and transfer learning on graphs, images, texts, and videos. It aims to fill the gaps between rich data sources, abundant machine learning libraries, and eager interdisciplinary researchers, with a focus on leveraging knowledge from multiple sources for accurate and interpretable prediction. To the best of our knowledge, this is the first publicly available Python library that considers both multimodal learning and transfer learning under a common framework of learning from multiple sources. It will make latest machine learning tools more accessible and accelerate the development of such tools. The name of the library consists of \textbf{Py} for \textbf{Py}thon, and \textbf{Kale} for \textbf{K}nowledge-\textbf{a}ware \textbf{le}arning. 

PyKale proposes a novel \textit{pipeline-based} application programming interface (API) to enforce standardization and minimalism, as shown in Figure \ref{pykale_pipeline}. It advocates our newly formulated \kalegreen{green machine learning} concepts of reducing repetitions and redundancy (Fig. \ref{fig_reduce}), reusing existing resources (Fig. \ref{fig_reuse}), and recycling learning models across areas (Fig. \ref{fig_recycle}) by building on standard software engineering practices, extending them, and tailoring the philosophies to machine learning. We include examples in bioinformatics, knowledge graph, image/video recognition, and medical imaging in PyKale. This library was motivated by needs in healthcare applications and thus considers healthcare as a primary domain of usage. PyKale is largely built on PyTorch and leverages many packages in the PyTorch ecosystem,\footnote{https://pytorch.org/ecosystem/} with the aim to become part of it. Our logo in Fig. \ref{pykale_logo} reflects the above characteristics, using an icon of simplified kale leaves.

Specifically, the main contributions are fourfold:
\begin{itemize}
    \item We introduce the PyKale library for knowledge-aware machine learning. It focuses on leveraging knowledge from multiple sources for accurate and interpretable prediction to enable and accelerate interdisciplinary research.
    \item We propose a pipeline-based API for a standardized and minimal design to help break interdisciplinary barriers. We advocate green machine learning concepts of reduce, reuse, and recycle in such a design.
    \item We demonstrate the usage of PyKale on real-world examples from multiple disciplines including bioinformatics, knowledge graph, image/video recognition, and medical imaging.
    \item We provide many community-engaging features including a detailed documentation, a project board to show the progress and road map, and GitHub discussions open to all users.  
\end{itemize}

The rest of this paper is organized as follows. We first review the state of the art open source software packages that are related to PyKale in Section \ref{sec_relatedwork}. Then we discuss the design principles and  API structure of PyKale in Section \ref{sec_design}. Next, we describe the usage of PyKale in Section \ref{sec_usage} and show example use cases from different applications in Section \ref{sec_examples}. Finally, we show the openness of our package and discuss its limitations and future developments in Section \ref{sec_openplan}, with conclusions drawn in Section \ref{sec_conclusion}.

The PyKale library is publicly available at \url{https://github.com/pykale/pykale} with accompanying data (mainly for testing at the moment) at \url{https://github.com/pykale/data} under an MIT license. PyKale can be installed from the Python Package Index (PyPI) via \codetxt{pip install pykale}. PyKale documentation is hosted at \url{https://pykale.readthedocs.io}. The primary targeted users are researchers and practitioners who have experience in Python and PyTorch programming and need to apply or develop machine learning systems taking data from multiple sources for prediction tasks, particularly in interdisciplinary areas such as healthcare. This paper refers to release version \textbf{0.1.0rc2}. 

\section{Related Work}\label{sec_relatedwork}
As an open source project, we have learned from numerous libraries in the public domain to build ours. Here, we can only briefly mention several that have been particularly influential or relevant. In particular, we focus on those PyTorch-based libraries that we have frequently studied in our development, regretfully omitting many libraries, such as those based on TensorFlow.

\subsection{PyTorch ecosystem}
PyTorch is a popular open source machine learning library, particularly for computer vision and NLP applications. The PyTorch ecosystem has a rich collection of tools and libraries for the development of advanced machine learning and AI systems. PyKale aims to fill the gap within the PyTorch ecosystem to support more interdisciplinary research based on multiple data sources. Therefore, we make extensive usage of existing libraries from the PyTorch ecosystem to reduce duplicated implementation.

\subsubsection{PyTorch Lightning}
PyTorch Lightning is a popular deep learning framework providing a high-level interface for PyTorch. By removing boilerplate code, it simplifies the development of research code and improves the reproducibility, flexibility,  and readability of the resulting models \cite{falcon2019pytorch}. The goal of PyKale shares some similarity with PyTorch Lightning, but with a different focus on supporting interdisciplinary research. We have lots of inspirations from the design of PyTorch Lightning.

\subsubsection{Other PyTorch libraries}
PyKale depends on some other libraries from the PyTorch ecosystem including TensorLy for tensor analysis \cite{JMLR:v20:18-277}, TorchVision for computer vision \cite{torch_vision}, and PyTorch Geometric for graph analysis \cite{pytorch_geometric}. We also learned from MONAI for medical image analysis \cite{nic_ma_2021_4891800}, GPyTorch for Gaussian processes \cite{GPyTorch}, Kornia for computer vision \cite{riba2020kornia}, and TorchIO for medical imaging preprocessing \cite{perez-garcia_torchio_2020}. These libraries have focuses different from ours on interdisciplinary research and multiple data sources.

The above libraries are listed as references at the end of our contributing guideline page.\footnote{https://github.com/pykale/pykale/blob/main/.github/CONTRIBUTING.md}

\subsection{Multimodal/transfer learning libraries}
To the best of our knowledge, no other libraries in the PyTorch ecosystem have the same pipeline-based API design as PyKale. Several PyTorch-based libraries on  multimodal learning or transfer learning are summarized below.

\subsubsection{MultiModal Framework (MMF)} MMF \cite{singh2020mmf} is the only library in the current PyTorch ecosystem focusing on multimodal learning of vision and language data in applications such as visual question answering, image captioning, visual dialog, hate detection and other vision and language tasks. PyKale differs from MMF not only in the API design, but also in the scientific fields covered and interdisciplinarity. PyKale aims to support interdisciplinary research such as medical imaging and drug discovery, and includes examples in these areas in the current release.

\subsubsection{Transfer-Learning-Library} The Transfer-Learning-Library \cite{dalib} is a library on transfer learning, providing domain adaptation and fine tuning algorithms for computer vision applications. PyKale differs not only in the API design but also in the multiple modalities of data supported, including also graphs and texts, as well as in interdisciplinarity.

\subsubsection{Cornac} Cornac \cite{salah2020cornac} is a library for multimodal recommender systems, leveraging auxiliary data (e.g., item descriptive text and image, social network, etc). PyKale differs from Cornac not only in the API design but also in the more diverse machine learning models and applications supported, not limited to recommender systems.

\subsubsection{ADA} ADA \cite{adalib2020}, with a full name ``\textit{(Yet) Another Domain Adaptation library}'', is an excellent package built on PyTorch Lightning for unsupervised and semi-supervised domain adaptation. We refactored ADA and made many changes for adaption to our pipeline-based API. We include a docstring at the top of each module adapted from ADA to indicate the source at ADA. Beyond a new API design, PyKale extends ADA substantially to support video domain adaptation and also supports non-vision data for interdisciplinary research.

There are some other smaller libraries on multimodal or transfer learning that are more narrowly focused than the above, such as Multimodal-Toolkit \cite{multimodal_toolkit}. PyKale frames multimodal learning and transfer learning under one roof of knowledge-aware machine learning from multiple sources with a unified pipeline-based API, aiming to support interdisciplinary research rather than just popular vision or language tasks.

\section{PyKale Design}\label{sec_design}

\subsection{Green machine learning}
\kalegreen{Green machine learning}  (and \kalegreen{green AI}) is a scarcely used term referring to energy-efficient computing  \cite{candelieri2021green,kung2014kernel,garcia2017energy}. Here, we  propose a new \textit{green machine learning} perspective for machine learning software development by formulating the \textbf{3R} guiding principles below. We build these principles on standard software engineering practices by extending them and tailoring the philosophies to machine learning:

\begin{itemize}
    \item \textbf{Reduce} repetition and redundancy
    \begin{itemize}
        \item Refactor code to standardize workflow and enforce styles, e.g., we refactored the deep drug-target binding affinity (DeepDTA) \cite{_zt_rk_2018} into our PyKale pipeline (Fig. \ref{fig_drugkaleapi}).
        \item Identify and remove duplicated functionalities, e.g., construct data loading API for popular dataset to share among different projects.
    \end{itemize} 
    \item \textbf{Reuse} existing resources
    \begin{itemize}
        \item Reuse the same machine learning pipeline for different data and applications, such as using the same multilinear principal component analysis (MPCA) pipeline for gait \cite{mpca}, brain \cite{Remurs}, and heart \cite{swift2021machine}.
        \item Reuse existing libraries (e.g., those in the PyTorch ecosystem, such as PyTorch Geometric) for available functionalities rather than implementing them again.
    \end{itemize} 
    \item \textbf{Recycle} learning models across areas
    \begin{itemize}
        \item Identify commonalities between applications, e.g., the similarity between commercial recommender systems (predicting user-item interactions) and drug discovery (predicting drug-target interactions).
        \item Recycle models for one application to another, e.g. from recommender systems \cite{bai2019joint} to drug discovery \cite{xu2020gripnet}.
    \end{itemize} 
\end{itemize}

Although the above is largely based on standard software engineering practices, this new formulation offers a new perspective to focus on core principles of standardization and minimalism. It has guided us to design a unique pipeline-based API to unify workflow, break barriers between areas and applications, and cross boundaries to fuse existing ideas and nurture new ideas. 

\subsection{Pipeline-based API}
Inspired by the convenience of machine learning pipelines in machine learning library like Spark MLlib \cite{meng2016mllib} and scikit-learn \cite{scikit-learn}, we design PyKale with a pipeline-based API as shown in Fig. \ref{fig_kaleapi}. This design has six key steps and embodies our green machine learning principles above by organizing code along a standardized machine learning pipeline to identify commonalities, reduce redundancy, and minimize cognitive overhead. 

In the following, we explain our unified API by starting with what the input and output are. We provide Python code snippets to help this explanation in Code \ref{code:sampleAPI}, mainly using the domain adaptation for digit classification example with a pipeline in Fig. \ref{fig_digitkaleapi}.\footnote{\url{https://github.com/pykale/pykale/tree/main/examples/digits_dann_lightn}} Figure \ref{fig_drugkaleapi} shows another pipeline for drug discovery. More code snippets are in Section \ref{sec_examples}.

\subsubsection{Load} The \codetxt{kale.loaddata} module mainly takes source paths (local or online) as the input and constructs dataloaders for datasets as the output. Its primary function is to load data for input to the machine learning system/pipeline. See line 2--7 of Code \ref{code:sampleAPI} for an example of loading digit images from multiple sources and line 8--13 of Code \ref{code:deepdta} for an example of loading drug and targets data.

\begin{code}[t]
\begin{minted}
[linenos,fontsize=\footnotesize,xleftmargin=2mm,numbersep=3pt,baselinestretch=1.2,frame=lines]{python}
# Load digits from multiple sources [digits_dann_lightn/main.py]
from kale.loaddata.digits_access import DigitDataset
from kale.loaddata.multi_domain import MultiDomainDatasets

source, target, _ = DigitDataset.get_source_target(
        DigitDataset("MNIST"), DigitDataset("USPS"), data_path)
dataset = MultiDomainDatasets(source, target))   

# Preprocess digits [kale/loaddata/digits_access.py]
import kale.prepdata.image_transform as image_transform

self._transform = image_transform.get_transform(transform_kind)
def get_train(self):
    return MNIST(data_path, train=True, transform=self._transform)

# Embed digit representations [digits_dann_lightn/model.py]
from kale.embed.image_cnn import SmallCNNFeature

# Predict digit class and domain [digits_dann_lightn/model.py]
from kale.predict.class_domain_nets import ClassNetSmallImage,
    DomainNetSmallImage

# Build domain adaption pipeline [digits_dann_lightn/model.py]
import kale.pipeline.domain_adapter as domain_adapter

model = domain_adapter.create_dann_based(method="DANN", 
            dataset=dataset,
            feature_extractor= SmallCNNFeature(),
            task_classifier=ClassNetSmallImage(),
            critic=DomainNetSmallImage(),
            **train_params)

# Utility functions [digits_dann_lightn/main.py] 
from kale.utils.csv_logger import setup_logger
from kale.utils.seed import set_seed
\end{minted}
\captionof{listing}{Code snippets from the source code for the digits domain adaptation example at \codetxt{pykale/examples/digits\_dann\_lightn/main.py} to demonstrate the unified pipeline-based API, simplified for inclusion here.} %
\label{code:sampleAPI}
\end{code}

\subsubsection{Preprocess} The \codetxt{kale.prepdata} module takes the loaded raw input data as input and preprocesses (transforms) them into a suitable representation for the following machine learning modules. Preprocessing steps include data normalization, augmentation, and other transformations of data representation not involving machine learning. Its submodules are typically imported in \codetxt{kale.loaddata}. See line 10--14 of Code \ref{code:sampleAPI} for an example of standardizing digit images with predefined transforms.

\subsubsection{Embed} The \codetxt{kale.embed} module takes preprocessed, normalized data representations to \textit{learn} new representations in a new space as the output. It includes dimensionality reduction algorithms (feature extraction and feature selection), such as MPCA \cite{mpca} and CNNs. They can be viewed as encoders or embedding functions that learn suitable representations from data. This is a machine learning module. See line 17 of Code \ref{code:sampleAPI} for an example of (importing) a CNN feature extractor.

\subsubsection{Predict} The \codetxt{kale.predict} module takes the learned (or preprocessed, if skipping \codetxt{kale.embed} ) representations to predict a desired target value as the output. Thus, this module provides prediction functions or decoders that learn a mapping from the input representation to a target prediction. This is also a machine learning module. See line 20--21 of Code \ref{code:sampleAPI} for an example of (importing) digit and domain classifiers.

\subsubsection{Evaluate} The \codetxt{kale.evaluate} module evaluates the prediction performance using some metrics. We reuse metrics from other libraries (e.g., \codetxt{sklearn.metrics} in line 4 of Code \ref{code:mpca-pipe}) and only implement metrics not commonly available, such as the Concordance Index (CI) \cite{ahuja2019joint} for measuring the proportion of concordant pairs. See line 19 of Code \ref{code:deepdta} for its example usage.

\subsubsection{Interpret} The \codetxt{kale.interpret} module aims to provide functions for interpretation of the learned features, the model, or the prediction results/outputs, e.g., via further analysis or visualization, and we only implement functions not commonly available. This module has implemented functions for selecting and visualizing weights from a trained model. See line 16--20 of Code \ref{code:mpca-pipe} for an example of visualizing weights of a linear model for interpretation.

\subsubsection{Pipeline} The \codetxt{kale.pipeline} module provides mature, off-the-shelf machine learning pipelines for ``plug-in usage''. Its submodules typically specify a machine learning workflow by combining several other modules. See line 24--31 of Code \ref{code:sampleAPI} for an example of calling a domain adaptation pipeline.

\subsubsection{Utilities} The \codetxt{kale.utils} module provides common utility functions, such as setting random seeds, logging results, or downloading data. See line 34--35 of Code \ref{code:sampleAPI} for examples of importing the seed-setting and csv-logging submodules. 

\subsection{Machine learning models}
Machine learning models in PyKale can be categorized into four main (possibly overlapping) groups.

\subsubsection{Multimodal learning} To support learning from data of multiple modalities, we need to first support learning from each individual modality. Thanks to the rich PyTorch ecosystem, we can build upon other libraries to have machine learning models supporting graphs, images, texts, and videos, primarily using PyTorch Dataloaders. The only missing major modality is audio but we have ongoing effort to include it in the near future, building upon \codetxt{torchaudio}.

Learning from heterogeneous data sources and data integration can be viewed as multimodal learning as well. To this end, PyKale has built a DeepDTA \cite{ztrk2018DeepDTADD} pipeline \codetxt{kale.pipeline.deep\_dti} that learns from drug and target data, the chemical representation of which can be transformed into sequence or vector representations. PyKale  also implemented our recent Graph information propagation Network (GripNet) \cite{xu2020gripnet} \codetxt{kale.embed.gripnet} for link prediction and data integration on heterogeneous knowledge graphs. PyKale has an  example \codetxt{drug\_gripnet} to show the model usage on public bioinformatics knowledge graph with drug and protein's information. %

\subsubsection{Transfer learning}
In transfer learning, PyKale currently focuses on domain adaptation \cite{ben2007analysis,pan2010domain}. We largely inherited the excellent, modular architecture from ADA \cite{adalib2020}, covering many important semi-supervised and unsupervised domain adaptation algorithms, such as domain-adversarial neural networks (DANN) \cite{ganin2016domain}, conditional adversarial domain adaptation networks (CDAN) \cite{long2018conditional}, deep adaptation networks (DAN) \cite{long2015learning} and joint adaptation networks (JAN) \cite{long2017deep}, and Wasserstein distance guided representation learning (WDGRL) \cite{shen2018wasserstein}. These algorithms are applicable to all modalities with appropriate representations. PyKale currently has two pipelines for domain adaptation: \codetxt{domain\_adapter} and \codetxt{video\_domain\_adapter} in \codetxt{kale.pipeline}.

\subsubsection{Deep learning}
PyKale builds deep neural networks (DNNs) upon the PyTorch API. Current implementations include CNNs \cite{Krizhevsky2012ImageNetCW} / 3D CNNs \cite{carreira2017quo, tran2018closer}, GCNs \cite{Kipf:2016tc}, and attention-based networks such as transformers \cite{vaswani2017attention} and squeeze and excitation networks  \cite{hu2018squeeze} (see more in Section \ref{sec_examples}). We use TorchVision \cite{torch_vision}, PyTorch Geometric \cite{pytorch_geometric}, and PyTorch Lightning \cite{falcon2019pytorch} in our implementation.

\subsubsection{Dimensionality reduction}
PyKale has built a Python version of the MPCA algorithm \cite{mpca} at \codetxt{kale.embed.mpca}, as well as an MPCA-based pipeline at \codetxt{kale.pipeline.mpca\_trainer}, using both the scikit-Learn library \cite{scikit-learn} and the TensorLy library \cite{JMLR:v20:18-277}. This pipeline has been successfully used for interpretable prediction in gait recognition from video sequences \cite{mpca}, cardiovascular disease diagnosis \cite{swift2021machine} and prognosis \cite{uthoff2020geodesically} using cardiac magnetic resonance imaging (MRI), and brain state classification using functional MRI (fMRI). We are further building into PyKale other advanced tensor-based %
algorithms such as regularized Multilinear Regression and Selection (Remurs) \cite{Remurs} and sparse tubal-regularized multilinear regression (Sturm) \cite{li2019sturm}.%

\subsection{Software engineering}

The PyKale team includes machine learning researchers and Research Software Engineers (RSEs). We have adopted good software engineering practices in a research context, often based on other libraries, particularly those in the PyTorch ecosystem.

\subsubsection{Version control and collaboration}

We use git for version control and GitHub for collaborative working. The PyKale repository stipulates a license (MIT) and contributing guidelines.\footnote{https://github.com/pykale/pykale/blob/main/.github/CONTRIBUTING.md} These enable reuse of the software and sustainability of the project through community contributions. This is a platform for long term availability of the resource and lays the foundation for community maintenance over an indefinite period.

\subsubsection{Documentation}

We use ``docstrings'' to embed documentation within the source code to maximize synchronicity between code and documentation. Sphinx\footnote{https://www.sphinx-doc.org/} is used to automatically build these (along with additional information in ``reStructuredText'' format) into html docs. Documentation is published via \url{readthedocs.com} and is kept up to date by Continuous Integration (CI). This is useful for keeping users and developers up-to-date with new features and bug fixes in a sustainable way. Detailed installation instructions are included.\footnote{https://pykale.readthedocs.io/en/latest/installation.html}

\subsubsection{Tests and continuous integration}

We use the PyTest framework and currently have 88\% test coverage. The test suite can be run locally, and also runs automatically on GitHub and must pass for code to be merged into the \codetxt{main} branch. This ensures that new features do not create unintended side-effects. CI is implemented using GitHub workflows/actions.\footnote{https://github.com/pykale/pykale/tree/main/.github/workflows} Our CI checks include static analysis, pre-commit checks (e.g., maximum file size), documentation building (via Read the Docs), changelog update, project assignment for issues and pull requests, PyPI release of packages, PyTest tests on multiple platforms and multiple python versions, and Codecov code coverage report.  This ensures that the version in the \codetxt{main} branch is always the most up to date working version, and meets our standards of functionality and coding style.

To maintain a small repository size (currently less than 1MB), we store test data in a separate repository at \url{https://github.com/pykale/data}, which can be downloaded via \codetxt{download\_file\_by\_url} in  \codetxt{kale.utils.download} automatically.

\section{PyKale Usage}\label{sec_usage}
Interdisciplinary research is a complex subject to support and care has to be taken to lower the barriers to entry. PyKale includes examples and tutorials to help users' exploration.

\subsection{Usage of pipeline-based API in examples}

PyKale examples are highly standardized. Each example typically has three essential modules (\codetxt{main.py}, \codetxt{config.py}, \codetxt{model.py}), one optional directory (\codetxt{configs}), and possibly other modules (\codetxt{trainer.py}):

\begin{itemize}
    \item \codetxt{main.py} is the main module to be run, showing the main workflow.
    \item \codetxt{config.py} is the configuration module that sets up the data, prediction problem, hyper-parameters, etc. The settings in this module are the default configuration.
    \item \codetxt{configs} is the directory to place customized configurations for individual runs. We use \codetxt{.yaml} files (see Section \ref{sec_yaml}) for this purpose.
    \item \codetxt{model.py} is the model module to define the machine learning model and configure its training parameters.
    \item  \codetxt{trainer.py} is the trainer module to define the training and testing workflow. This module is only needed when NOT using PyTorch Lightning.
\end{itemize}

\subsection{Building new modules or projects}
Users can build new modules or projects following the steps below.
\begin{itemize}
    \item Step 1 - Examples: Choose one of the examples of the users' interest (e.g., most relevant to the users' project) to
    \begin{itemize}
        \item browse through the configuration, main, and model modules,
        \item download the data if needed, and
        \item run the example following instructions in the example's README.
    \end{itemize}
    \item Step 2a - New model: To develop new machine learning models under PyKale,
    \begin{itemize}
        \item define the blocks in the users' pipeline to figure out what the methods are for data loading, preprocessing data, embedding (encoder/representation), prediction (decoder), evaluation, and interpretation, and
        \item modify existing pipelines with the users' customized blocks or build a new pipeline with \codetxt{pykale} blocks and blocks from other libraries.
    \end{itemize}
    \item Step 2b - New applications: To develop new applications using PyKale,
    \begin{itemize}
        \item clarify the input data and the prediction target to find matching functionalities in \codetxt{pykale} (request if not found), and
        \item tailor data loading, preprocessing, and evaluation (and interpretation if needed) to the users' application.
    \end{itemize}
\end{itemize}

\begin{code}[t]
\begin{minted}
[linenos,fontsize=\footnotesize,xleftmargin=2mm,numbersep=3pt,baselinestretch=1.2,frame=lines]{python}
# The file config.py that defines the default configuration
from yacs.config import CfgNode as CN

# Config definition
_C = CN()

# Dataset
_C.DATASET = CN()
_C.DATASET.ROOT = "../data"
_C.DATASET.NAME = "CIFAR10"

# Solver
_C.SOLVER = CN()
_C.SOLVER.SEED = 2020
_C.SOLVER.BASE_LR = 0.05
_C.SOLVER.TRAIN_BATCH_SIZE = 128
_C.SOLVER.MAX_EPOCHS = 100

# ISONet configs
_C.ISON = CN()
_C.ISON.DEPTH = 34

# Misc options
_C.OUTPUT_DIR = "./outputs"
\end{minted}
\begin{minted}
[linenos,fontsize=\footnotesize,xleftmargin=2mm,numbersep=3pt,baselinestretch=1.2,frame=lines]{yaml}
# Customization in a .yaml file
SOLVER:
  BASE_LR: 0.01
  MAX_EPOCHS: 10  # For quick testing
ISON:
  DEPTH: 38 
\end{minted}

\captionof{listing}{Code snippets to demonstrate the usage of YAML to configure machine learning systems in PyKale, from \codetxt{pykale/examples/cifar\_isonet/config.py}, which is adapted from \url{https://github.com/HaozhiQi/ISONet}.}
\label{code:yaml}
\end{code}

\subsection{YAML configuration}\label{sec_yaml} 
PyKale examples configure a machine learning system using YAML \cite{ben2009yaml}. This is inspired by the usage of YAML in the GitHub package for the Isometric Network (ISONet) \cite{qi2020deep},\footnote{https://github.com/HaozhiQi/ISONet} with our adapted version illustrated in Code \ref{code:yaml}. As modern machine learning systems typically have many settings to configure, specifying many/all settings in command line or Python modules becomes difficult to manage and read. Using YAML greatly improves the readability and reproducibility, and makes configuration changes much easier, via a default configuration specified in \codetxt{config.py} (top of Code \ref{code:yaml}) and customized configurations specified in a respective \codetxt{.yaml} file (bottom of Code \ref{code:yaml}), which will be merged to overwrite the default setting at run time.

\subsection{Notebook tutorials with Binder and Colab}\label{sec_tutorial}

We have eight real-world examples of PyKale usage.\footnote{https://github.com/pykale/pykale/tree/main/examples} However, tutorials without the need of any installation are important for new users to get familiar with the PyKale workflow and API. For these we must scale-back on real-world datasets due to the computational resources needed, as these lead to long runtimes, unsuitable for interactive learning. Therefore, we are simplifying our examples into Jupyter notebook tutorials so that each tutorial takes minutes instead of hours to run. This will strike a balance between computational requirements and runtime, without resorting to toy examples.

\begin{figure}[!t]
  \centering
  \subfigure[Binder]{	\label{fig_binder}
		\includegraphics[width=0.8\linewidth]{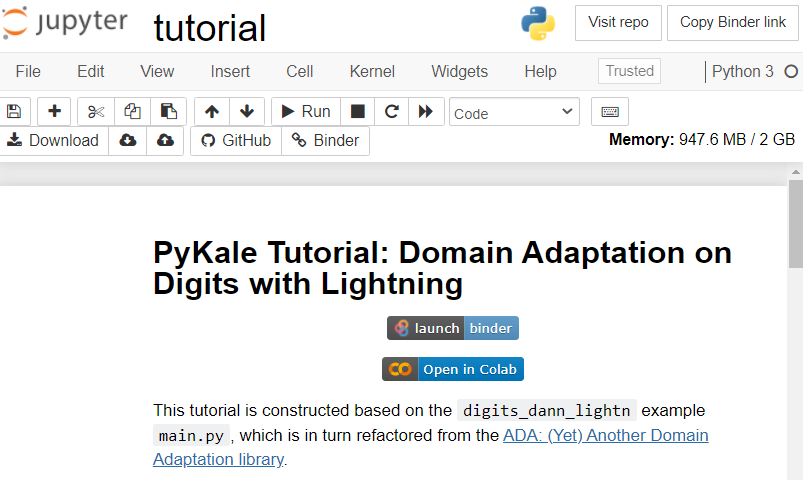}
  }
  \subfigure[Google Colab]{	\label{fig_colab}
		\includegraphics[width=0.8\linewidth]{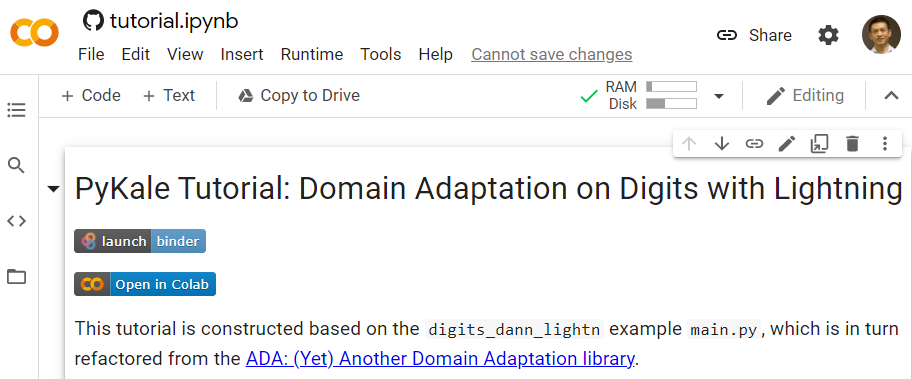}
  }  
  \caption{PyKale digits domain adaptation example on cloud-based services.}\label{pykale_tutorial}
\end{figure}

To bring further convenience, we set up cloud-based services with both Binder\footnote{https://mybinder.org/} and Google Colaboratory (Colab)\footnote{https://colab.research.google.com/} for our notebook tutorials so that any users can run PyKale tutorials without the need of any installation. The first such tutorial has been released,\footnote{\url{https://github.com/pykale/pykale/blob/main/examples/digits_dann_lightn/tutorial.ipynb}} with screenshots in Figure \ref{pykale_tutorial}. More such tutorials are in development.

\section{Use Cases: PyKale Examples}\label{sec_examples}
PyKale currently has example applications from three areas below:
\begin{itemize}
    \item Image/video recognition: classification of images (objects, digits) or videos (actions in first-person videos);
    \item Bioinformatics/graph analysis: prediction of links between entities in knowledge graphs (BindingDB, BioSNAP-Decagon);
    \item Medical imaging: disease diagnosis from cardiac MRIs.
\end{itemize}

The above examples deal with graphs, images, and videos. Our current APIs support text processing (e.g., for NLP tasks) but an example in this area is still in development. We are also conducting research in integrating audio features in action recognition so an example involving audio data is a future task as well. Examples in computer vision applications such as image and video recognition are a good start for most users due to the popularity of vision applications and a low barrier to entry (e.g., no need for specific domain knowledge as in drug discovery). Models first developed in computer vision can be reused or recycled for other applications.

The data used in PyKale examples are real-world data frequently used in research papers. Subsequently, it may take quite some time to finish running these examples. For quick running and demonstration of the workflow, tutorials (Section \ref{sec_tutorial}) are simplified examples that serve as a better starting point. The following subsections give an overview of the data and algorithms used in the example machine learning systems in PyKale.

\subsection{CIFAR and digits classification}
Small-image datasets are good for building examples of real-world relevance. PyKale has three such examples: two on the CIFAR datasets \cite{cifar-10} and one on digits datasets including MNIST \cite{lecun1998mnist,deng2012mnist}, modified MNIST \cite{ganin2016domain}, USPS \cite{hull1994database}, and SVHN \cite{shvndataset}. \codetxt{cifar\_isonet} is the first example in PyKale refactoring the ISONet code \url{https://github.com/HaozhiQi/ISONet} on CIFAR10 and CIFAR100 into the PyKale API. \codetxt{cifar\_cnntransformer} is another example on CIFAR showing the simple application of a mixed CNN and transformer model for vision tasks. Code \ref{code:sampleAPI} has shown code snippets of digits classification via domain adaptation using MNIST as the source domain and USPS as the target domain, in the \codetxt{digits\_dann\_lightn} example. 

Examples on these popular datasets could help users familiar with them make easy connections between the PyKale API and what they are already familiar with.

\begin{code}[t]
\begin{minted}
[linenos,fontsize=\footnotesize,xleftmargin=2mm,numbersep=3pt,baselinestretch=1.2,frame=lines]{python}
# Transform and dataset for multi-modal video data
from kale.prepdata.video_transform import get_transform
from kale.loaddata.video_datasets import BasicVideoDataset

transform = get_transform(transform_kind, image_modality)
dataset = BasicVideoDataset(data_path, train_list, 
                            imagefile_template="frame_{:010d}.jpg"
                            if image_modality in ["rgb"] 
                            else "flow_{}_{:010d}.jpg",
                            transform=transform,
                            image_modality=image_modality)

# Action video feature extractor
from kale.embed.video_feature_extractor import
                                          get_video_feat_extractor
feature_extractor = get_video_feat_extractor("I3D", image_modality,
                                           attention, num_classes)

\end{minted}
\captionof{listing}{Code snippets to demonstrate the example on action recognition via domain adaptation.} 
\label{code:ardaAPI}
\end{code}

\subsection{Action recognition via domain adaptation}
\subsubsection{Data}
For this action recognition example, we constructed three first-person vision datasets, ADL$_{small}$, GTEA-KITCHEN and EPIC$_{cvpr20}$, with 222/454/10094 action videos respectively. We selected and reorganized three videos from ADL dataset \cite{pirsiavash2012detecting} as three domains of ADL$_{small}$, re-annotated GTEA \cite{li2015delving, fathi2011learning} and KITCHEN \cite{de2009guide} datasets to build GTEA-KITCHEN, and adopted the public dataset EPIC$_{cvpr20}$ \cite{munro20multi}. We will provide instructions on how to construct these datasets from the source at \url{https://github.com/pykale/data/tree/main/video_data/video_test_data} (in progress). Each dataset has two modalities: RGB and optical flow. %

\subsubsection{Algorithms}
For video feature extraction, we built two state-of-the-art action recognition algorithms, I3D \cite{carreira2017quo} and 3D ResNet \cite{tran2018closer}, into PyKale. For domain adaptation, we extended the domain adaptation framework for images (digits, adapted from \cite{adalib2020}) to videos. We followed the same pipeline as digits classification while providing additional specific functions for action videos and multi-modal data. As shown in Code \ref{code:ardaAPI}, the image modality parameter can be set to choose the proper data transform and loader for different modalities: RGB, optical flow, and joint, and all video feature extractors are accessed via a unified interface.

\subsection{Drug-target interaction prediction}
\subsubsection{Data}
Predicting the binding affinity between drug compounds and target proteins is fundamental for drug discovery and drug repurposing. This example uses three public datasets (for three metrics Ki, Kd and IC50) from BindingDB \cite{Liu2007BindingDBAW} containing 52,284, 375,032, and 991,486 interaction pairs respectively, accessed via the Therapeutics Data Commons (TDC) platform \cite{tdc}. Given the amino acid sequences of targets and SMILES (Simplified Molecular Input Line Entry System) strings of drug compounds, the task is to predict drug-target binding affinity. Following \cite{Karimi2019DeepAffinityID}, the affinity metrics are transformed into the logarithm form for more stable training and validation.

\subsubsection{Algorithms}
For this drug-target interaction prediction problem, we built DeepDTA \cite{ztrk2018DeepDTADD} into PyKale, a typical CNN-based model with encoder-decoder architecture. The drug SMILES string and target amino acid sequence are encoded by their independent CNNs, and then a multilayer perceptron is used to decode the affinity from the drug-target combined encoding. We refactored DeepDTA into the PyKale API structure with separate modules for load data, preprocess data, encode (embed) and decode (predict). These refactored modules are flexible and can be reused across different applications. Code \ref{code:deepdta} illustrates the usage of the DeepDTA model as implemented in PyKale.

\begin{code}[t]
\begin{minted}
[linenos,fontsize=\footnotesize,xleftmargin=2mm,numbersep=3pt,baselinestretch=1.2,frame=lines]{python}
from kale.loaddata.tdc_datasets import BindingDBDataset
from kale.embed.seq_nn import CNNEncoder
from kale.predict.decode import MLPDecoder
from kale.pipeline.deep_dti import DeepDTATrainer
import pytorch_lightning as pl

# Load training, validation, and test data
train_dataset = BindingDBDataset(name="Kd", split="train")
val_dataset = BindingDBDataset(name="Kd", split="valid")
test_dataset = BindingDBDataset(name="Kd", split="valid")
train_loader = DataLoader(dataset=train_dataset, batch_size=64)
val_loader = DataLoader(dataset=val_dataset, batch_size=64)
test_loader = DataLoader(dataset=test_dataset, batch_size=64)

# Initialize DeepDTA with encoder and decoder
drug_encoder, target_encoder = CNNEncoder(), CNNEncoder()
decoder = MLPDecoder()
model = DeepDTATrainer(drug_encoder, target_encoder, decoder, 
                       ci_metric=True)

# Train the model under the PyTorch Lightning framework
trainer = pl.Trainer(max_epochs=100)
trainer.fit(model, train_dataloader=train_loader,
            val_dataloaders=val_loader)
trainer.test(test_dataloaders=test_loader)
\end{minted}
\captionof{listing}{Demonstration code for drug-target interaction prediction with DeepDTA.}
\label{code:deepdta}
\end{code}

\subsection{Polypharmacy side effect prediction via knowledge graph link prediction}

\subsubsection{Data}
Polypharmacy uses drug combination to treat complex diseases, which may cause side effects. Predicting such side effects can be formulated as a link prediction problem on knowledge graphs of drugs and proteins \cite{Novek2020PredictingPS, Zitnik2018ModelingPS, xu2020gripnet}. This example uses the public BioSNAP-Decagon dataset \cite{biosnapnets} with 6,075,428 edges, 1,100 different edge labels, and three types of edges: drug-drug interaction, protein-protein interaction and drug-target interaction. The drug-drug interactions model polypharmacy side effects.

\subsubsection{Algorithms}

We built our recent GripNet \cite{xu2020gripnet} into PyKale for this example. GripNet is a subgraph network framework for multi-relational link prediction on heterogeneous graphs via segregated node representation learning on ``supervertices'' and ``superedges''. PyKale implements APIs to support these advanced concepts for node embedding and link prediction on heterogeneous graphs. %

\begin{code}[t]
\begin{minted}
[linenos,fontsize=\footnotesize,xleftmargin=2mm,numbersep=3pt,baselinestretch=1.2,frame=lines]{python}
from kale.pipeline.mpca_trainer import MPCATrainer
from kale.interpret import model_weights
from sklearn.model_selection import train_test_split
from sklearn.metrics import accuracy_score, roc_auc_score

# Assume data x and labels y are loaded and preprocessed
trainer = MPCATrainer(mpca_params={"return_vector": True})
x_train, x_test, y_train, y_test = train_test_split(x, y, 
                                                    test_size=0.2)
trainer.fit(x_train, y_train)
print("Accuracy:", accuracy_score(y_test, trainer.predict(x_test))
print("AUC:", roc_auc_score(y_test, 
                            trainer.decision_function(x_test))

# Visualize model coefficients (weights) for interpretation
weights = trainer.mpca.inverse_transform(trainer.clf.coef_) 
          - trainer.mpca.mean_
top_weights = model_weights.select_top_weight(weights)
model_weights.plot_weights(top_weights[0][0], 
                           background_img=x[0][0])

\end{minted}
\captionof{listing}{MPCA-based pipeline for predicting pulmonary arterial hypertension with cardiac MRI.}
\label{code:mpca-pipe}
\end{code}

\subsection{Cardiac MRI classification}
\subsubsection{Data}
The data used to build this example are a subset of the dataset used in \cite{swift2021machine}, which consists of cardiac MRI (CMRI) sequences acquired from patients with pulmonary arterial hypertension and health controls. They are not yet in the public domain but we are exploring release options. The CMRI sequences are standardized using methods in \codetxt{kale.prepdata.image\_transform}. %

\subsubsection{Algorithms}
This example uses the \codetxt{MPCATrainer} pipeline in \codetxt{kale.pipeline.mpca\_trainer}, which implemented the machine learning pipeline used in \cite{swift2021machine,song2015learning,mpca} in the scikit-learn style. The three key steps are MPCA dimensionality reduction, feature selection, and classification (SVM, Linear SVM, or logisitic regression), where the feature selection and classification algorithms reuse respective APIs in scikit-learn \cite{scikit-learn}. Code \ref{code:mpca-pipe} shows the steps for MPCA-based prediction and interpretation on cardiac MRI.

\section{PyKale Openness and Plan}\label{sec_openplan}

\subsection{License and community engagement}
PyKale is publicly available at \url{https://github.com/pykale/pykale} under an MIT license, which is a simple permissive license with minimal restrictions. The PyKale GitHub repository has active discussion board at \url{https://github.com/pykale/pykale/discussions} and project board at \url{https://github.com/pykale/pykale/projects} to interact with users and make the development process and plan transparent to all users. Complete documentation is hosted at \url{https://pykale.readthedocs.io/} with multiple versions available, generated automatically from \url{https://github.com/pykale/pykale/tree/main/docs}. We have provided tutorials and examples as well as detailed contributing guidelines at \url{https://github.com/pykale/pykale/blob/main/.github/CONTRIBUTING.md} and change logs at \url{https://github.com/pykale/pykale/blob/main/.github/CHANGELOG.md}. 

We also released a 12-minute YouTube video at \url{https://youtu.be/i5BYdMfbpMQ} to briefly explain the motivation and principles behind PyKale. This paper is another effort to reach out to the wider research community to share this resource and get feedback for further improvements.

\subsection{Limitations and future development}
PyKale is an open-source project started in June 2020, with the first PyPI release in January 2021. It was motivated by the growing needs for machine learning systems that can deal with multiple sources of data, particularly in interdisciplinary areas such as healthcare. For example, clinicians often need to make use of a combination of medical images (e.g., X-rays, CTs, MRIs), biological data (gene, DNA, RNA), and electronic health record for decision making. Our focus on multimodal learning and transfer learning has defined a challenging scope, while holding the promises to break barriers in interdisciplinary research. 

To date, PyKale has built APIs supporting machine learning from graphs, images, texts, and videos, with four mature pipelines implemented. Nevertheless, we do not have an example on text data yet, and we have not built APIs for audio yet. Developing projects involving multiple data sources takes considerably longer time than developing those involving a single data source. The current version of PyKale has two examples on multimodal learning involving heterogeneous drug and target data and two examples on domain adaptation (transfer learning) for images and videos. These examples laid solid foundations for us to grow our research  in these areas and build more advanced examples in future development. In addition, our tests currently have a coverage of 88\%. We need further improvements for a higher coverage and more rigorous tests.

\section{Conclusions}\label{sec_conclusion}
In this paper, we have introduced \textit{PyKale}, a Python library for knowledge-aware machine learning from multiple sources, particularly from multiple modalities for multimodal learning and from multiple domains for transfer learning. This library was motivated by needs in healthcare applications (hence the acronym kale, a healthy vegetable) and aims to enable and accelerate interdisciplinary research. Building on standard software engineering practices, we proposed a new green machine learning perspective to advocate reducing repetitions and redundancy, reusing existing resources, and recycling learning models across areas. Following such principles, we designed our API to be pipeline-based to unify the workflow and increase the flexibility. This design can help to break barriers between different areas or applications and facilitate the fusion and nurture of ideas across discipline boundaries. 

The goal of PyKale is to facilitate interdisciplinary, knowledge-aware machine learning research for graphs, images, texts, and videos. It will make it easier to bring machine learning models developed in one area to the other, and integrate data from multiple sources for prediction tasks in interdisciplinary areas. Its focus on leveraging knowledge from multiple sources also helps accurate and interpretable prediction. 
To demonstrate such potential, we have shown example applications including bioinformatics, knowledge graph, image/video recognition, and medical imaging on real-world datasets.

\begin{acks}
The development of PyKale is partially supported by the Innovator Awards: Digital Technologies from the Wellcome Trust (grant 215799/Z/19/Z). We thank the support and contributions from David Jones, Will Furnass, and other members of the Research Software Engineering (RSE) team headed by Paul Richmond. We also thank early users of our library and their helpful feedback. %
\end{acks}

\bibliographystyle{ACM-Reference-Format}
\bibliography{pykale}


\begin{thebibliography}{64}


\ifx \showCODEN    \undefined \def \showCODEN     #1{\unskip}     \fi
\ifx \showDOI      \undefined \def \showDOI       #1{#1}\fi
\ifx \showISBNx    \undefined \def \showISBNx     #1{\unskip}     \fi
\ifx \showISBNxiii \undefined \def \showISBNxiii  #1{\unskip}     \fi
\ifx \showISSN     \undefined \def \showISSN      #1{\unskip}     \fi
\ifx \showLCCN     \undefined \def \showLCCN      #1{\unskip}     \fi
\ifx \shownote     \undefined \def \shownote      #1{#1}          \fi
\ifx \showarticletitle \undefined \def \showarticletitle #1{#1}   \fi
\ifx \showURL      \undefined \def \showURL       {\relax}        \fi
\providecommand\bibfield[2]{#2}
\providecommand\bibinfo[2]{#2}
\providecommand\natexlab[1]{#1}
\providecommand\showeprint[2][]{arXiv:#2}

\bibitem[\protect\citeauthoryear{Ahuja and van~der Schaar}{Ahuja and van~der
  Schaar}{2019}]%
        {ahuja2019joint}
\bibfield{author}{\bibinfo{person}{Kartik Ahuja} {and} \bibinfo{person}{Mihaela
  van~der Schaar}.} \bibinfo{year}{2019}\natexlab{}.
\newblock \showarticletitle{Joint Concordance Index}. In
  \bibinfo{booktitle}{\emph{Proceedings of the 2019 53rd Asilomar Conference on
  Signals, Systems, and Computers}}. \bibinfo{pages}{2206--2213}.
\newblock


\bibitem[\protect\citeauthoryear{Bai, Ge, Liu, and Lu}{Bai
  et~al\mbox{.}}{2019}]%
        {bai2019joint}
\bibfield{author}{\bibinfo{person}{Peizhen Bai}, \bibinfo{person}{Yan Ge},
  \bibinfo{person}{Fangling Liu}, {and} \bibinfo{person}{Haiping Lu}.}
  \bibinfo{year}{2019}\natexlab{}.
\newblock \showarticletitle{Joint interaction with context operation for
  collaborative filtering}.
\newblock \bibinfo{journal}{\emph{Pattern Recognition}}  \bibinfo{volume}{88}
  (\bibinfo{year}{2019}), \bibinfo{pages}{729--738}.
\newblock


\bibitem[\protect\citeauthoryear{Ben-David, Blitzer, Crammer, Pereira,
  et~al\mbox{.}}{Ben-David et~al\mbox{.}}{2007}]%
        {ben2007analysis}
\bibfield{author}{\bibinfo{person}{Shai Ben-David}, \bibinfo{person}{John
  Blitzer}, \bibinfo{person}{Koby Crammer}, \bibinfo{person}{Fernando Pereira},
  {et~al\mbox{.}}} \bibinfo{year}{2007}\natexlab{}.
\newblock \showarticletitle{Analysis of representations for domain adaptation}.
  In \bibinfo{booktitle}{\emph{Proceedings of the Advances in Neural
  Information Processing Systems}}. \bibinfo{pages}{137--144}.
\newblock


\bibitem[\protect\citeauthoryear{Ben-Kiki, Evans, and Ingerson}{Ben-Kiki
  et~al\mbox{.}}{2009}]%
        {ben2009yaml}
\bibfield{author}{\bibinfo{person}{Oren Ben-Kiki}, \bibinfo{person}{Clark
  Evans}, {and} \bibinfo{person}{Brian Ingerson}.}
  \bibinfo{year}{2009}\natexlab{}.
\newblock \showarticletitle{Yaml ain't markup language (yaml™) version 1.1}.
\newblock \bibinfo{journal}{\emph{Working Draft 2008-05}}  \bibinfo{volume}{11}
  (\bibinfo{year}{2009}).
\newblock


\bibitem[\protect\citeauthoryear{Candelieri, Perego, and Archetti}{Candelieri
  et~al\mbox{.}}{2021}]%
        {candelieri2021green}
\bibfield{author}{\bibinfo{person}{Antonio Candelieri},
  \bibinfo{person}{Riccardo Perego}, {and} \bibinfo{person}{Francesco
  Archetti}.} \bibinfo{year}{2021}\natexlab{}.
\newblock \showarticletitle{Green machine learning via augmented Gaussian
  processes and multi-information source optimization}.
\newblock \bibinfo{journal}{\emph{Soft Computing}} (\bibinfo{year}{2021}),
  \bibinfo{pages}{1--13}.
\newblock


\bibitem[\protect\citeauthoryear{Carion, Massa, Synnaeve, Usunier, Kirillov,
  and Zagoruyko}{Carion et~al\mbox{.}}{2020}]%
        {carion2020end}
\bibfield{author}{\bibinfo{person}{Nicolas Carion}, \bibinfo{person}{Francisco
  Massa}, \bibinfo{person}{Gabriel Synnaeve}, \bibinfo{person}{Nicolas
  Usunier}, \bibinfo{person}{Alexander Kirillov}, {and} \bibinfo{person}{Sergey
  Zagoruyko}.} \bibinfo{year}{2020}\natexlab{}.
\newblock \showarticletitle{End-to-end object detection with transformers}. In
  \bibinfo{booktitle}{\emph{Proceedings of the European Conference on Computer
  Vision}}. \bibinfo{pages}{213--229}.
\newblock


\bibitem[\protect\citeauthoryear{Carreira and Zisserman}{Carreira and
  Zisserman}{2017}]%
        {carreira2017quo}
\bibfield{author}{\bibinfo{person}{Joao Carreira} {and} \bibinfo{person}{Andrew
  Zisserman}.} \bibinfo{year}{2017}\natexlab{}.
\newblock \showarticletitle{Quo vadis, action recognition? a new model and the
  kinetics dataset}. In \bibinfo{booktitle}{\emph{Proceedings of the IEEE
  Conference on Computer Vision and Pattern Recognition}}.
  \bibinfo{pages}{6299--6308}.
\newblock


\bibitem[\protect\citeauthoryear{De~la Torre, Hodgins, Bargteil, Martin, Macey,
  Collado, and Beltran}{De~la Torre et~al\mbox{.}}{2009}]%
        {de2009guide}
\bibfield{author}{\bibinfo{person}{Fernando De~la Torre},
  \bibinfo{person}{Jessica Hodgins}, \bibinfo{person}{Adam Bargteil},
  \bibinfo{person}{Xavier Martin}, \bibinfo{person}{Justin Macey},
  \bibinfo{person}{Alex Collado}, {and} \bibinfo{person}{Pep Beltran}.}
  \bibinfo{year}{2009}\natexlab{}.
\newblock \showarticletitle{Guide to the carnegie mellon university multimodal
  activity (cmu-mmac) database}.
\newblock  (\bibinfo{year}{2009}).
\newblock


\bibitem[\protect\citeauthoryear{Deng}{Deng}{2012}]%
        {deng2012mnist}
\bibfield{author}{\bibinfo{person}{Li Deng}.} \bibinfo{year}{2012}\natexlab{}.
\newblock \showarticletitle{The mnist database of handwritten digit images for
  machine learning research}.
\newblock \bibinfo{journal}{\emph{IEEE Signal Processing Magazine}}
  \bibinfo{volume}{29}, \bibinfo{number}{6} (\bibinfo{year}{2012}),
  \bibinfo{pages}{141--142}.
\newblock


\bibitem[\protect\citeauthoryear{Dosovitskiy, Beyer, Kolesnikov, and
  et~al}{Dosovitskiy et~al\mbox{.}}{2021}]%
        {dosovitskiy2021an}
\bibfield{author}{\bibinfo{person}{Alexey Dosovitskiy}, \bibinfo{person}{Lucas
  Beyer}, \bibinfo{person}{Alexander Kolesnikov}, {and} \bibinfo{person}{et
  al}.} \bibinfo{year}{2021}\natexlab{}.
\newblock \showarticletitle{An Image is Worth 16x16 Words: Transformers for
  Image Recognition at Scale}. In \bibinfo{booktitle}{\emph{Proceedings of the
  International Conference on Learning Representations}}.
\newblock


\bibitem[\protect\citeauthoryear{Falcon and et~al.}{Falcon and et~al.}{2019}]%
        {falcon2019pytorch}
\bibfield{author}{\bibinfo{person}{William~A Falcon} {and} \bibinfo{person}{et
  al.}} \bibinfo{year}{2019}\natexlab{}.
\newblock \showarticletitle{PyTorch Lightning}.
\newblock \bibinfo{journal}{\emph{GitHub.}}  \bibinfo{volume}{3}
  (\bibinfo{year}{2019}).
\newblock
\urldef\tempurl%
\url{https://github.com/PyTorchLightning/pytorch-lightning}
\showURL{%
\tempurl}


\bibitem[\protect\citeauthoryear{Fathi, Ren, and Rehg}{Fathi
  et~al\mbox{.}}{2011}]%
        {fathi2011learning}
\bibfield{author}{\bibinfo{person}{Alireza Fathi}, \bibinfo{person}{Xiaofeng
  Ren}, {and} \bibinfo{person}{James~M Rehg}.} \bibinfo{year}{2011}\natexlab{}.
\newblock \showarticletitle{Learning to recognize objects in egocentric
  activities}. In \bibinfo{booktitle}{\emph{Proceedings of the IEEE Conference
  on Computer Vision and Pattern Recognition}}. \bibinfo{pages}{3281--3288}.
\newblock


\bibitem[\protect\citeauthoryear{Fey and Lenssen}{Fey and Lenssen}{2019}]%
        {pytorch_geometric}
\bibfield{author}{\bibinfo{person}{Matthias Fey} {and} \bibinfo{person}{Jan~E.
  Lenssen}.} \bibinfo{year}{2019}\natexlab{}.
\newblock \showarticletitle{Fast Graph Representation Learning with {PyTorch
  Geometric}}. In \bibinfo{booktitle}{\emph{ICLR Workshop on Representation
  Learning on Graphs and Manifolds}}.
\newblock


\bibitem[\protect\citeauthoryear{Ganin, Ustinova, Ajakan, Germain, Larochelle,
  Laviolette, Marchand, and Lempitsky}{Ganin et~al\mbox{.}}{2016}]%
        {ganin2016domain}
\bibfield{author}{\bibinfo{person}{Yaroslav Ganin}, \bibinfo{person}{Evgeniya
  Ustinova}, \bibinfo{person}{Hana Ajakan}, \bibinfo{person}{Pascal Germain},
  \bibinfo{person}{Hugo Larochelle}, \bibinfo{person}{Fran{\c{c}}ois
  Laviolette}, \bibinfo{person}{Mario Marchand}, {and} \bibinfo{person}{Victor
  Lempitsky}.} \bibinfo{year}{2016}\natexlab{}.
\newblock \showarticletitle{Domain-adversarial training of neural networks}.
\newblock \bibinfo{journal}{\emph{Journal of Machine Learning Research}}
  \bibinfo{volume}{17}, \bibinfo{number}{1} (\bibinfo{year}{2016}),
  \bibinfo{pages}{2096--2030}.
\newblock


\bibitem[\protect\citeauthoryear{Garc{\'\i}a~Mart{\'\i}n}{Garc{\'\i}a~Mart{\'\i}n}{2017}]%
        {garcia2017energy}
\bibfield{author}{\bibinfo{person}{Eva Garc{\'\i}a~Mart{\'\i}n}.}
  \bibinfo{year}{2017}\natexlab{}.
\newblock \showarticletitle{Energy efficiency in machine learning: A position
  paper}. In \bibinfo{booktitle}{\emph{Proceedings of the 30th Annual Workshop
  of the Swedish Artificial Intelligence Society}}, Vol.~\bibinfo{volume}{137}.
  \bibinfo{pages}{68--72}.
\newblock


\bibitem[\protect\citeauthoryear{Gardner, Pleiss, Weinberger, Bindel, and
  Wilson}{Gardner et~al\mbox{.}}{2018}]%
        {GPyTorch}
\bibfield{author}{\bibinfo{person}{Jacob Gardner}, \bibinfo{person}{Geoff
  Pleiss}, \bibinfo{person}{Kilian~Q Weinberger}, \bibinfo{person}{David
  Bindel}, {and} \bibinfo{person}{Andrew~G Wilson}.}
  \bibinfo{year}{2018}\natexlab{}.
\newblock \showarticletitle{GPyTorch: Blackbox Matrix-Matrix Gaussian Process
  Inference with GPU Acceleration}. In \bibinfo{booktitle}{\emph{Proceedings of
  the Advances in Neural Information Processing Systems}},
  Vol.~\bibinfo{volume}{31}. \bibinfo{pages}{7587–7597}.
\newblock


\bibitem[\protect\citeauthoryear{Georgian}{Georgian}{2020}]%
        {multimodal_toolkit}
\bibfield{author}{\bibinfo{person}{Georgian}.} \bibinfo{year}{2020}\natexlab{}.
\newblock \showarticletitle{Multimodal-Toolkit}.
\newblock \bibinfo{journal}{\emph{GitHub}} (\bibinfo{year}{2020}).
\newblock
\urldef\tempurl%
\url{https://github.com/georgian-io/Multimodal-Toolkit}
\showURL{%
\tempurl}


\bibitem[\protect\citeauthoryear{Hu, Shen, and Sun}{Hu et~al\mbox{.}}{2018}]%
        {hu2018squeeze}
\bibfield{author}{\bibinfo{person}{Jie Hu}, \bibinfo{person}{Li Shen}, {and}
  \bibinfo{person}{Gang Sun}.} \bibinfo{year}{2018}\natexlab{}.
\newblock \showarticletitle{Squeeze-and-excitation networks}. In
  \bibinfo{booktitle}{\emph{Proceedings of the IEEE Conference on Computer
  Vision and Pattern Recognition}}. \bibinfo{pages}{7132--7141}.
\newblock


\bibitem[\protect\citeauthoryear{Huang, Fu, Gao, and et~al}{Huang
  et~al\mbox{.}}{2021}]%
        {tdc}
\bibfield{author}{\bibinfo{person}{Kexin Huang}, \bibinfo{person}{Tianfan Fu},
  \bibinfo{person}{Wenhao Gao}, {and} \bibinfo{person}{et al}.}
  \bibinfo{year}{2021}\natexlab{}.
\newblock \showarticletitle{Therapeutics Data Commons: Machine Learning
  Datasets and Tasks for Therapeutics}.
\newblock \bibinfo{journal}{\emph{arXiv preprint arXiv:2102.09548}}
  (\bibinfo{year}{2021}).
\newblock


\bibitem[\protect\citeauthoryear{Hull}{Hull}{1994}]%
        {hull1994database}
\bibfield{author}{\bibinfo{person}{Jonathan Hull}.}
  \bibinfo{year}{1994}\natexlab{}.
\newblock \showarticletitle{A database for handwritten text recognition
  research}.
\newblock \bibinfo{journal}{\emph{IEEE Transactions on Pattern Analysis and
  Machine Intelligence}} \bibinfo{volume}{16}, \bibinfo{number}{5}
  (\bibinfo{year}{1994}), \bibinfo{pages}{550--554}.
\newblock


\bibitem[\protect\citeauthoryear{Jiang, Fu, and Long}{Jiang
  et~al\mbox{.}}{2020}]%
        {dalib}
\bibfield{author}{\bibinfo{person}{Junguang Jiang}, \bibinfo{person}{Bo Fu},
  {and} \bibinfo{person}{Mingsheng Long}.} \bibinfo{year}{2020}\natexlab{}.
\newblock \showarticletitle{Transfer-Learning-library}.
\newblock \bibinfo{journal}{\emph{GitHub}} (\bibinfo{year}{2020}).
\newblock
\urldef\tempurl%
\url{https://github.com/thuml/Transfer-Learning-Library}
\showURL{%
\tempurl}


\bibitem[\protect\citeauthoryear{Karimi, Wu, Wang, and Shen}{Karimi
  et~al\mbox{.}}{2019}]%
        {Karimi2019DeepAffinityID}
\bibfield{author}{\bibinfo{person}{Mostafa Karimi}, \bibinfo{person}{Di Wu},
  \bibinfo{person}{Zhangyang Wang}, {and} \bibinfo{person}{Yang Shen}.}
  \bibinfo{year}{2019}\natexlab{}.
\newblock \showarticletitle{DeepAffinity: interpretable deep learning of
  compound--protein affinity through unified recurrent and convolutional neural
  networks}.
\newblock \bibinfo{journal}{\emph{Bioinformatics}} \bibinfo{volume}{35},
  \bibinfo{number}{18} (\bibinfo{year}{2019}), \bibinfo{pages}{3329--3338}.
\newblock


\bibitem[\protect\citeauthoryear{Kipf and Welling}{Kipf and Welling}{2017}]%
        {Kipf:2016tc}
\bibfield{author}{\bibinfo{person}{Thomas Kipf} {and} \bibinfo{person}{Max
  Welling}.} \bibinfo{year}{2017}\natexlab{}.
\newblock \showarticletitle{Semi-Supervised Classification with Graph
  Convolutional Networks}. In \bibinfo{booktitle}{\emph{Proceedings of the 5th
  International Conference on Learning Representations}}.
\newblock


\bibitem[\protect\citeauthoryear{Kossaifi, Panagakis, Anandkumar, and
  Pantic}{Kossaifi et~al\mbox{.}}{2019}]%
        {JMLR:v20:18-277}
\bibfield{author}{\bibinfo{person}{Jean Kossaifi}, \bibinfo{person}{Yannis
  Panagakis}, \bibinfo{person}{Anima Anandkumar}, {and} \bibinfo{person}{Maja
  Pantic}.} \bibinfo{year}{2019}\natexlab{}.
\newblock \showarticletitle{TensorLy: Tensor Learning in Python}.
\newblock \bibinfo{journal}{\emph{Journal of Machine Learning Research}}
  \bibinfo{volume}{20}, \bibinfo{number}{26} (\bibinfo{year}{2019}),
  \bibinfo{pages}{1--6}.
\newblock


\bibitem[\protect\citeauthoryear{Krizhevsky, Nair, and Hinton}{Krizhevsky
  et~al\mbox{.}}{2010}]%
        {cifar-10}
\bibfield{author}{\bibinfo{person}{Alex Krizhevsky}, \bibinfo{person}{Vinod
  Nair}, {and} \bibinfo{person}{Geoffrey Hinton}.}
  \bibinfo{year}{2010}\natexlab{}.
\newblock \showarticletitle{Cifar-10 (canadian institute for advanced
  research)}.
\newblock   \bibinfo{volume}{5} (\bibinfo{year}{2010}).
\newblock


\bibitem[\protect\citeauthoryear{Krizhevsky, Sutskever, and Hinton}{Krizhevsky
  et~al\mbox{.}}{2012}]%
        {Krizhevsky2012ImageNetCW}
\bibfield{author}{\bibinfo{person}{Alex Krizhevsky}, \bibinfo{person}{Ilya
  Sutskever}, {and} \bibinfo{person}{Geoffrey~E Hinton}.}
  \bibinfo{year}{2012}\natexlab{}.
\newblock \showarticletitle{Imagenet classification with deep convolutional
  neural networks}. In \bibinfo{booktitle}{\emph{Proceedings of the Advances in
  Neural Information Processing Systems}}. \bibinfo{pages}{1097--1105}.
\newblock


\bibitem[\protect\citeauthoryear{Kung}{Kung}{2014}]%
        {kung2014kernel}
\bibfield{author}{\bibinfo{person}{Sun~Yuan Kung}.}
  \bibinfo{year}{2014}\natexlab{}.
\newblock \bibinfo{booktitle}{\emph{Kernel methods and machine learning}}.
\newblock \bibinfo{publisher}{Cambridge University Press}.
\newblock


\bibitem[\protect\citeauthoryear{LeCun}{LeCun}{1998}]%
        {lecun1998mnist}
\bibfield{author}{\bibinfo{person}{Yann LeCun}.}
  \bibinfo{year}{1998}\natexlab{}.
\newblock \showarticletitle{The MNIST database of handwritten digits}.
\newblock \bibinfo{journal}{\emph{http://yann. lecun. com/exdb/mnist/}}
  (\bibinfo{year}{1998}).
\newblock


\bibitem[\protect\citeauthoryear{Li, Lou, Zhou, and Lu}{Li
  et~al\mbox{.}}{2019}]%
        {li2019sturm}
\bibfield{author}{\bibinfo{person}{Wenwen Li}, \bibinfo{person}{Jian Lou},
  \bibinfo{person}{Shuo Zhou}, {and} \bibinfo{person}{Haiping Lu}.}
  \bibinfo{year}{2019}\natexlab{}.
\newblock \showarticletitle{Sturm: Sparse tubal-regularized multilinear
  regression for fmri}. In \bibinfo{booktitle}{\emph{Proceedings of the
  International Workshop on Machine Learning in Medical Imaging}}.
  \bibinfo{pages}{256--264}.
\newblock


\bibitem[\protect\citeauthoryear{Li, Ye, and Rehg}{Li et~al\mbox{.}}{2015}]%
        {li2015delving}
\bibfield{author}{\bibinfo{person}{Yin Li}, \bibinfo{person}{Zhefan Ye}, {and}
  \bibinfo{person}{James~M Rehg}.} \bibinfo{year}{2015}\natexlab{}.
\newblock \showarticletitle{Delving into egocentric actions}. In
  \bibinfo{booktitle}{\emph{Proceedings of the IEEE Conference on Computer
  Vision and Pattern Recognition}}. \bibinfo{pages}{287--295}.
\newblock


\bibitem[\protect\citeauthoryear{Liu, Lin, Wen, Jorissen, and Gilson}{Liu
  et~al\mbox{.}}{2007}]%
        {Liu2007BindingDBAW}
\bibfield{author}{\bibinfo{person}{Tiqing Liu}, \bibinfo{person}{Yuhmei Lin},
  \bibinfo{person}{Xin Wen}, \bibinfo{person}{R. Jorissen}, {and}
  \bibinfo{person}{M. Gilson}.} \bibinfo{year}{2007}\natexlab{}.
\newblock \showarticletitle{BindingDB: a web-accessible database of
  experimentally determined protein–ligand binding affinities}.
\newblock \bibinfo{journal}{\emph{Nucleic Acids Research}}
  \bibinfo{volume}{35} (\bibinfo{year}{2007}), \bibinfo{pages}{D198 -- D201}.
\newblock


\bibitem[\protect\citeauthoryear{Long, Cao, Wang, and Jordan}{Long
  et~al\mbox{.}}{2015}]%
        {long2015learning}
\bibfield{author}{\bibinfo{person}{Mingsheng Long}, \bibinfo{person}{Yue Cao},
  \bibinfo{person}{Jianmin Wang}, {and} \bibinfo{person}{Michael Jordan}.}
  \bibinfo{year}{2015}\natexlab{}.
\newblock \showarticletitle{Learning transferable features with deep adaptation
  networks}. In \bibinfo{booktitle}{\emph{Proceedings of the International
  Conference on Machine Learning}}. \bibinfo{pages}{97--105}.
\newblock


\bibitem[\protect\citeauthoryear{Long, CAO, Wang, and Jordan}{Long
  et~al\mbox{.}}{2018}]%
        {long2018conditional}
\bibfield{author}{\bibinfo{person}{Mingsheng Long}, \bibinfo{person}{ZHANGJIE
  CAO}, \bibinfo{person}{Jianmin Wang}, {and} \bibinfo{person}{Michael~I
  Jordan}.} \bibinfo{year}{2018}\natexlab{}.
\newblock \showarticletitle{Conditional Adversarial Domain Adaptation}. In
  \bibinfo{booktitle}{\emph{Proceedings of the Advances in Neural Information
  Processing Systems}}, Vol.~\bibinfo{volume}{31}.
\newblock


\bibitem[\protect\citeauthoryear{Long, Zhu, Wang, and Jordan}{Long
  et~al\mbox{.}}{2017}]%
        {long2017deep}
\bibfield{author}{\bibinfo{person}{Mingsheng Long}, \bibinfo{person}{Han Zhu},
  \bibinfo{person}{Jianmin Wang}, {and} \bibinfo{person}{Michael~I Jordan}.}
  \bibinfo{year}{2017}\natexlab{}.
\newblock \showarticletitle{Deep transfer learning with joint adaptation
  networks}. In \bibinfo{booktitle}{\emph{Proceedings of the International
  Conference on Machine Learning}}. \bibinfo{pages}{2208--2217}.
\newblock


\bibitem[\protect\citeauthoryear{Lu, Plataniotis, and Venetsanopoulos}{Lu
  et~al\mbox{.}}{2008}]%
        {mpca}
\bibfield{author}{\bibinfo{person}{Haiping Lu}, \bibinfo{person}{Konstantinos~N
  Plataniotis}, {and} \bibinfo{person}{Anastasios~N Venetsanopoulos}.}
  \bibinfo{year}{2008}\natexlab{}.
\newblock \showarticletitle{MPCA: Multilinear principal component analysis of
  tensor objects}.
\newblock \bibinfo{journal}{\emph{IEEE Transactions on Neural Networks}}
  \bibinfo{volume}{19}, \bibinfo{number}{1} (\bibinfo{year}{2008}),
  \bibinfo{pages}{18--39}.
\newblock


\bibitem[\protect\citeauthoryear{Ma, Li, and Brown}{Ma et~al\mbox{.}}{2021}]%
        {nic_ma_2021_4891800}
\bibfield{author}{\bibinfo{person}{Nic Ma}, \bibinfo{person}{Wenqi Li}, {and}
  \bibinfo{person}{Richard Brown}.} \bibinfo{year}{2021}\natexlab{}.
\newblock \bibinfo{booktitle}{\emph{Project-MONAI/MONAI: 0.5.3}}.
\newblock
\urldef\tempurl%
\url{https://doi.org/10.5281/zenodo.4891800}
\showDOI{\tempurl}


\bibitem[\protect\citeauthoryear{Marcel and Rodriguez}{Marcel and
  Rodriguez}{2010}]%
        {torch_vision}
\bibfield{author}{\bibinfo{person}{S\'{e}bastien Marcel} {and}
  \bibinfo{person}{Yann Rodriguez}.} \bibinfo{year}{2010}\natexlab{}.
\newblock \showarticletitle{Torchvision the Machine-Vision Package of Torch}.
  In \bibinfo{booktitle}{\emph{Proceedings of the 18th ACM International
  Conference on Multimedia}}. \bibinfo{pages}{1485–1488}.
\newblock


\bibitem[\protect\citeauthoryear{Meng, Bradley, Yavuz, Sparks, Venkataraman,
  Liu, Freeman, Tsai, Amde, Owen, et~al\mbox{.}}{Meng et~al\mbox{.}}{2016}]%
        {meng2016mllib}
\bibfield{author}{\bibinfo{person}{Xiangrui Meng}, \bibinfo{person}{Joseph
  Bradley}, \bibinfo{person}{Burak Yavuz}, \bibinfo{person}{Evan Sparks},
  \bibinfo{person}{Shivaram Venkataraman}, \bibinfo{person}{Davies Liu},
  \bibinfo{person}{Jeremy Freeman}, \bibinfo{person}{DB Tsai},
  \bibinfo{person}{Manish Amde}, \bibinfo{person}{Sean Owen}, {et~al\mbox{.}}}
  \bibinfo{year}{2016}\natexlab{}.
\newblock \showarticletitle{Mllib: Machine learning in apache spark}.
\newblock \bibinfo{journal}{\emph{Journal of Machine Learning Research}}
  \bibinfo{volume}{17}, \bibinfo{number}{1} (\bibinfo{year}{2016}),
  \bibinfo{pages}{1235--1241}.
\newblock


\bibitem[\protect\citeauthoryear{Munro and Damen}{Munro and Damen}{2020}]%
        {munro20multi}
\bibfield{author}{\bibinfo{person}{Jonathan Munro} {and} \bibinfo{person}{Dima
  Damen}.} \bibinfo{year}{2020}\natexlab{}.
\newblock \showarticletitle{Multi-modal domain adaptation for fine-grained
  action recognition}. In \bibinfo{booktitle}{\emph{Proceedings of the IEEE
  Conference on Computer Vision and Pattern Recognition}}.
  \bibinfo{pages}{122--132}.
\newblock


\bibitem[\protect\citeauthoryear{Netzer, Wang, Coates, Bissacco, Wu, and
  Ng}{Netzer et~al\mbox{.}}{2011}]%
        {shvndataset}
\bibfield{author}{\bibinfo{person}{Yuval Netzer}, \bibinfo{person}{Tao Wang},
  \bibinfo{person}{Adam Coates}, \bibinfo{person}{Alessandro Bissacco},
  \bibinfo{person}{Bo Wu}, {and} \bibinfo{person}{Andrew~Y. Ng}.}
  \bibinfo{year}{2011}\natexlab{}.
\newblock \showarticletitle{Reading Digits in Natural Images with Unsupervised
  Feature Learning}.
\newblock \bibinfo{journal}{\emph{NeurIPS Workshop on Deep Learning and
  Unsupervised Feature Learning}} (\bibinfo{year}{2011}).
\newblock


\bibitem[\protect\citeauthoryear{Nov{\'a}{\v{c}}ek and
  Mohamed}{Nov{\'a}{\v{c}}ek and Mohamed}{2020}]%
        {Novek2020PredictingPS}
\bibfield{author}{\bibinfo{person}{V{\'\i}t Nov{\'a}{\v{c}}ek} {and}
  \bibinfo{person}{Sameh~K Mohamed}.} \bibinfo{year}{2020}\natexlab{}.
\newblock \showarticletitle{Predicting Polypharmacy Side-effects Using
  Knowledge Graph Embeddings.}. In \bibinfo{booktitle}{\emph{Proceedings of the
  AMIA Joint Summits on Translational Science}}. \bibinfo{pages}{449--458}.
\newblock


\bibitem[\protect\citeauthoryear{{\"O}zt{\"u}rk, Olmez, and
  {\"O}zg{\"u}r}{{\"O}zt{\"u}rk et~al\mbox{.}}{2018}]%
        {ztrk2018DeepDTADD}
\bibfield{author}{\bibinfo{person}{Hakime {\"O}zt{\"u}rk}, \bibinfo{person}{E.
  Olmez}, {and} \bibinfo{person}{Arzucan {\"O}zg{\"u}r}.}
  \bibinfo{year}{2018}\natexlab{}.
\newblock \showarticletitle{DeepDTA: deep drug–target binding affinity
  prediction}.
\newblock \bibinfo{journal}{\emph{Bioinformatics}}  \bibinfo{volume}{34}
  (\bibinfo{year}{2018}), \bibinfo{pages}{i821 -- i829}.
\newblock


\bibitem[\protect\citeauthoryear{Pan, Tsang, Kwok, and Yang}{Pan
  et~al\mbox{.}}{2010}]%
        {pan2010domain}
\bibfield{author}{\bibinfo{person}{Sinno~Jialin Pan}, \bibinfo{person}{Ivor~W
  Tsang}, \bibinfo{person}{James~T Kwok}, {and} \bibinfo{person}{Qiang Yang}.}
  \bibinfo{year}{2010}\natexlab{}.
\newblock \showarticletitle{Domain adaptation via transfer component analysis}.
\newblock \bibinfo{journal}{\emph{IEEE Transactions on Neural Networks}}
  \bibinfo{volume}{22}, \bibinfo{number}{2} (\bibinfo{year}{2010}),
  \bibinfo{pages}{199--210}.
\newblock


\bibitem[\protect\citeauthoryear{Pedregosa, Varoquaux, Gramfort, Michel,
  Thirion, Grisel, Blondel, Prettenhofer, Weiss, Dubourg, Vanderplas, Passos,
  Cournapeau, Brucher, Perrot, and Duchesnay}{Pedregosa et~al\mbox{.}}{2011}]%
        {scikit-learn}
\bibfield{author}{\bibinfo{person}{F. Pedregosa}, \bibinfo{person}{G.
  Varoquaux}, \bibinfo{person}{A. Gramfort}, \bibinfo{person}{V. Michel},
  \bibinfo{person}{B. Thirion}, \bibinfo{person}{O. Grisel},
  \bibinfo{person}{M. Blondel}, \bibinfo{person}{P. Prettenhofer},
  \bibinfo{person}{R. Weiss}, \bibinfo{person}{V. Dubourg}, \bibinfo{person}{J.
  Vanderplas}, \bibinfo{person}{A. Passos}, \bibinfo{person}{D. Cournapeau},
  \bibinfo{person}{M. Brucher}, \bibinfo{person}{M. Perrot}, {and}
  \bibinfo{person}{E. Duchesnay}.} \bibinfo{year}{2011}\natexlab{}.
\newblock \showarticletitle{Scikit-learn: Machine Learning in {P}ython}.
\newblock \bibinfo{journal}{\emph{Journal of Machine Learning Research}}
  \bibinfo{volume}{12} (\bibinfo{year}{2011}), \bibinfo{pages}{2825--2830}.
\newblock


\bibitem[\protect\citeauthoryear{P{\'e}rez-Garc{\'i}a, Sparks, and
  Ourselin}{P{\'e}rez-Garc{\'i}a et~al\mbox{.}}{2020}]%
        {perez-garcia_torchio_2020}
\bibfield{author}{\bibinfo{person}{Fernando P{\'e}rez-Garc{\'i}a},
  \bibinfo{person}{Rachel Sparks}, {and} \bibinfo{person}{Sebastien Ourselin}.}
  \bibinfo{year}{2020}\natexlab{}.
\newblock \showarticletitle{{TorchIO}: a {Python} library for efficient
  loading, preprocessing, augmentation and patch-based sampling of medical
  images in deep learning}.
\newblock  (\bibinfo{year}{2020}).
\newblock
\urldef\tempurl%
\url{http://arxiv.org/abs/2003.04696}
\showURL{%
\tempurl}


\bibitem[\protect\citeauthoryear{Pirsiavash and Ramanan}{Pirsiavash and
  Ramanan}{2012}]%
        {pirsiavash2012detecting}
\bibfield{author}{\bibinfo{person}{Hamed Pirsiavash} {and}
  \bibinfo{person}{Deva Ramanan}.} \bibinfo{year}{2012}\natexlab{}.
\newblock \showarticletitle{Detecting activities of daily living in
  first-person camera views}. In \bibinfo{booktitle}{\emph{Proceedings of the
  IEEE Conference on Computer Vision and Pattern Recognition}}.
  \bibinfo{pages}{2847--2854}.
\newblock


\bibitem[\protect\citeauthoryear{Qi, You, Wang, Ma, and Malik}{Qi
  et~al\mbox{.}}{2020}]%
        {qi2020deep}
\bibfield{author}{\bibinfo{person}{Haozhi Qi}, \bibinfo{person}{Chong You},
  \bibinfo{person}{Xiaolong Wang}, \bibinfo{person}{Yi Ma}, {and}
  \bibinfo{person}{Jitendra Malik}.} \bibinfo{year}{2020}\natexlab{}.
\newblock \showarticletitle{Deep isometric learning for visual recognition}. In
  \bibinfo{booktitle}{\emph{Proceedings of the International Conference on
  Machine Learning}}. \bibinfo{pages}{7824--7835}.
\newblock


\bibitem[\protect\citeauthoryear{Riba, Mishkin, Ponsa, Rublee, and
  Bradski}{Riba et~al\mbox{.}}{2020}]%
        {riba2020kornia}
\bibfield{author}{\bibinfo{person}{Edgar Riba}, \bibinfo{person}{Dmytro
  Mishkin}, \bibinfo{person}{Daniel Ponsa}, \bibinfo{person}{Ethan Rublee},
  {and} \bibinfo{person}{Gary Bradski}.} \bibinfo{year}{2020}\natexlab{}.
\newblock \showarticletitle{Kornia: an open source differentiable computer
  vision library for pytorch}. In \bibinfo{booktitle}{\emph{Proceedings of the
  IEEE Winter Conference on Applications of Computer Vision}}.
  \bibinfo{pages}{3674--3683}.
\newblock


\bibitem[\protect\citeauthoryear{Ronneberger, Fischer, and Brox}{Ronneberger
  et~al\mbox{.}}{2015}]%
        {ronneberger2015unet}
\bibfield{author}{\bibinfo{person}{Olaf Ronneberger}, \bibinfo{person}{Philipp
  Fischer}, {and} \bibinfo{person}{Thomas Brox}.}
  \bibinfo{year}{2015}\natexlab{}.
\newblock \showarticletitle{U-net: Convolutional networks for biomedical image
  segmentation}. In \bibinfo{booktitle}{\emph{Proceedings of the International
  Conference on Medical Image Computing and Computer-Assisted Intervention}}.
  \bibinfo{pages}{234--241}.
\newblock


\bibitem[\protect\citeauthoryear{Salah, Truong, and Lauw}{Salah
  et~al\mbox{.}}{2020}]%
        {salah2020cornac}
\bibfield{author}{\bibinfo{person}{Aghiles Salah}, \bibinfo{person}{Quoc-Tuan
  Truong}, {and} \bibinfo{person}{Hady~W Lauw}.}
  \bibinfo{year}{2020}\natexlab{}.
\newblock \showarticletitle{Cornac: A Comparative Framework for Multimodal
  Recommender Systems}.
\newblock \bibinfo{journal}{\emph{Journal of Machine Learning Research}}
  \bibinfo{volume}{21}, \bibinfo{number}{95} (\bibinfo{year}{2020}),
  \bibinfo{pages}{1--5}.
\newblock


\bibitem[\protect\citeauthoryear{Shen, Qu, Zhang, and Yu}{Shen
  et~al\mbox{.}}{2018}]%
        {shen2018wasserstein}
\bibfield{author}{\bibinfo{person}{Jian Shen}, \bibinfo{person}{Yanru Qu},
  \bibinfo{person}{Weinan Zhang}, {and} \bibinfo{person}{Yong Yu}.}
  \bibinfo{year}{2018}\natexlab{}.
\newblock \showarticletitle{Wasserstein distance guided representation learning
  for domain adaptation}. In \bibinfo{booktitle}{\emph{Proceedings of the AAAI
  Conference on Artificial Intelligence}}, Vol.~\bibinfo{volume}{32}.
\newblock


\bibitem[\protect\citeauthoryear{Singh, Goswami, Natarajan, Jiang, Chen, Shah,
  Rohrbach, Batra, and Parikh}{Singh et~al\mbox{.}}{2020}]%
        {singh2020mmf}
\bibfield{author}{\bibinfo{person}{Amanpreet Singh}, \bibinfo{person}{Vedanuj
  Goswami}, \bibinfo{person}{Vivek Natarajan}, \bibinfo{person}{Yu Jiang},
  \bibinfo{person}{Xinlei Chen}, \bibinfo{person}{Meet Shah},
  \bibinfo{person}{Marcus Rohrbach}, \bibinfo{person}{Dhruv Batra}, {and}
  \bibinfo{person}{Devi Parikh}.} \bibinfo{year}{2020}\natexlab{}.
\newblock \bibinfo{title}{MMF: A multimodal framework for vision and language
  research}.
\newblock
  \bibinfo{howpublished}{\url{https://github.com/facebookresearch/mmf}}.
\newblock


\bibitem[\protect\citeauthoryear{Song and Lu}{Song and Lu}{2017}]%
        {Remurs}
\bibfield{author}{\bibinfo{person}{Xiaonan Song} {and} \bibinfo{person}{Haiping
  Lu}.} \bibinfo{year}{2017}\natexlab{}.
\newblock \showarticletitle{Multilinear regression for embedded feature
  selection with application to fmri analysis}. In
  \bibinfo{booktitle}{\emph{Proceedings of the AAAI Conference on Artificial
  Intelligence}}, Vol.~\bibinfo{volume}{31}.
\newblock


\bibitem[\protect\citeauthoryear{Song, Meng, Shi, and Lu}{Song
  et~al\mbox{.}}{2015}]%
        {song2015learning}
\bibfield{author}{\bibinfo{person}{Xiaonan Song}, \bibinfo{person}{Lingnan
  Meng}, \bibinfo{person}{Qiquan Shi}, {and} \bibinfo{person}{Haiping Lu}.}
  \bibinfo{year}{2015}\natexlab{}.
\newblock \showarticletitle{Learning tensor-based features for whole-brain fMRI
  classification}. In \bibinfo{booktitle}{\emph{Proceedings of the
  International Conference on Medical Image Computing and Computer-Assisted
  Intervention}}. \bibinfo{pages}{613--620}.
\newblock


\bibitem[\protect\citeauthoryear{Swift, Lu, Uthoff, Garg, Cogliano, Taylor,
  Metherall, Zhou, Johns, Alabed, et~al\mbox{.}}{Swift et~al\mbox{.}}{2021}]%
        {swift2021machine}
\bibfield{author}{\bibinfo{person}{Andrew~J Swift}, \bibinfo{person}{Haiping
  Lu}, \bibinfo{person}{Johanna Uthoff}, \bibinfo{person}{Pankaj Garg},
  \bibinfo{person}{Marcella Cogliano}, \bibinfo{person}{Jonathan Taylor},
  \bibinfo{person}{Peter Metherall}, \bibinfo{person}{Shuo Zhou},
  \bibinfo{person}{Christopher~S Johns}, \bibinfo{person}{Samer Alabed},
  {et~al\mbox{.}}} \bibinfo{year}{2021}\natexlab{}.
\newblock \showarticletitle{A machine learning cardiac magnetic resonance
  approach to extract disease features and automate pulmonary arterial
  hypertension diagnosis}.
\newblock \bibinfo{journal}{\emph{European Heart Journal-Cardiovascular
  Imaging}} \bibinfo{volume}{22}, \bibinfo{number}{2} (\bibinfo{year}{2021}),
  \bibinfo{pages}{236--245}.
\newblock


\bibitem[\protect\citeauthoryear{Tousch and Renaudin}{Tousch and
  Renaudin}{2020}]%
        {adalib2020}
\bibfield{author}{\bibinfo{person}{Anne-Marie Tousch} {and}
  \bibinfo{person}{Christophe Renaudin}.} \bibinfo{year}{2020}\natexlab{}.
\newblock \bibinfo{title}{(Yet) Another Domain Adaptation library}.
\newblock
\newblock
\urldef\tempurl%
\url{https://github.com/criteo-research/pytorch-ada}
\showURL{%
\tempurl}


\bibitem[\protect\citeauthoryear{Tran, Wang, Torresani, Ray, LeCun, and
  Paluri}{Tran et~al\mbox{.}}{2018}]%
        {tran2018closer}
\bibfield{author}{\bibinfo{person}{Du Tran}, \bibinfo{person}{Heng Wang},
  \bibinfo{person}{Lorenzo Torresani}, \bibinfo{person}{Jamie Ray},
  \bibinfo{person}{Yann LeCun}, {and} \bibinfo{person}{Manohar Paluri}.}
  \bibinfo{year}{2018}\natexlab{}.
\newblock \showarticletitle{A closer look at spatiotemporal convolutions for
  action recognition}. In \bibinfo{booktitle}{\emph{Proceedings of the IEEE
  Conference on Computer Vision and Pattern Recognition}}.
  \bibinfo{pages}{6450--6459}.
\newblock


\bibitem[\protect\citeauthoryear{Uthoff, Alabed, Swift, and Lu}{Uthoff
  et~al\mbox{.}}{2020}]%
        {uthoff2020geodesically}
\bibfield{author}{\bibinfo{person}{Johanna Uthoff}, \bibinfo{person}{Samer
  Alabed}, \bibinfo{person}{Andrew~J Swift}, {and} \bibinfo{person}{Haiping
  Lu}.} \bibinfo{year}{2020}\natexlab{}.
\newblock \showarticletitle{Geodesically Smoothed Tensor Features for Pulmonary
  Hypertension Prognosis Using the Heart and Surrounding Tissues}. In
  \bibinfo{booktitle}{\emph{Proceedings of the International Conference on
  Medical Image Computing and Computer-Assisted Intervention}}.
  \bibinfo{pages}{253--262}.
\newblock


\bibitem[\protect\citeauthoryear{Vaswani, Shazeer, Parmar, Uszkoreit, Jones,
  Gomez, Kaiser, and Polosukhin}{Vaswani et~al\mbox{.}}{2017}]%
        {vaswani2017attention}
\bibfield{author}{\bibinfo{person}{Ashish Vaswani}, \bibinfo{person}{Noam
  Shazeer}, \bibinfo{person}{Niki Parmar}, \bibinfo{person}{Jakob Uszkoreit},
  \bibinfo{person}{Llion Jones}, \bibinfo{person}{Aidan Gomez},
  \bibinfo{person}{Ukasz Kaiser}, {and} \bibinfo{person}{Illia Polosukhin}.}
  \bibinfo{year}{2017}\natexlab{}.
\newblock \showarticletitle{Attention is All You Need}. In
  \bibinfo{booktitle}{\emph{Proceedings of the Advances in Neural Information
  Processing Systems}}. \bibinfo{pages}{6000–6010}.
\newblock


\bibitem[\protect\citeauthoryear{Wang, Girshick, Gupta, and He}{Wang
  et~al\mbox{.}}{2018}]%
        {wang2018non}
\bibfield{author}{\bibinfo{person}{Xiaolong Wang}, \bibinfo{person}{Ross
  Girshick}, \bibinfo{person}{Abhinav Gupta}, {and} \bibinfo{person}{Kaiming
  He}.} \bibinfo{year}{2018}\natexlab{}.
\newblock \showarticletitle{Non-local neural networks}. In
  \bibinfo{booktitle}{\emph{Proceedings of the IEEE Conference on Computer
  Vision and Pattern Recognition}}. \bibinfo{pages}{7794--7803}.
\newblock


\bibitem[\protect\citeauthoryear{Xu, Sang, Bai, Yang, and Lu}{Xu
  et~al\mbox{.}}{2020}]%
        {xu2020gripnet}
\bibfield{author}{\bibinfo{person}{Hao Xu}, \bibinfo{person}{Shengqi Sang},
  \bibinfo{person}{Peizhen Bai}, \bibinfo{person}{Laurence Yang}, {and}
  \bibinfo{person}{Haiping Lu}.} \bibinfo{year}{2020}\natexlab{}.
\newblock \bibinfo{title}{GripNet: Graph Information Propagation on Supergraph
  for Heterogeneous Graphs}.
\newblock
\newblock
\showeprint[arxiv]{2010.15914}~[cs.LG]


\bibitem[\protect\citeauthoryear{Zitnik, Agrawal, and Leskovec}{Zitnik
  et~al\mbox{.}}{2018a}]%
        {Zitnik2018ModelingPS}
\bibfield{author}{\bibinfo{person}{M. Zitnik}, \bibinfo{person}{Monica
  Agrawal}, {and} \bibinfo{person}{J. Leskovec}.}
  \bibinfo{year}{2018}\natexlab{a}.
\newblock \showarticletitle{Modeling polypharmacy side effects with graph
  convolutional networks}.
\newblock \bibinfo{journal}{\emph{Bioinformatics}}  \bibinfo{volume}{34}
  (\bibinfo{year}{2018}), \bibinfo{pages}{i457 -- i466}.
\newblock


\bibitem[\protect\citeauthoryear{Zitnik, Sosi\v{c}, Maheshwari, and
  Leskovec}{Zitnik et~al\mbox{.}}{2018b}]%
        {biosnapnets}
\bibfield{author}{\bibinfo{person}{Marinka Zitnik}, \bibinfo{person}{Rok
  Sosi\v{c}}, \bibinfo{person}{Sagar Maheshwari}, {and} \bibinfo{person}{Jure
  Leskovec}.} \bibinfo{year}{2018}\natexlab{b}.
\newblock \bibinfo{title}{{BioSNAP Datasets}: {Stanford} Biomedical Network
  Dataset Collection}.
\newblock
\newblock


\bibitem[\protect\citeauthoryear{Öztürk, Özgür, and Ozkirimli}{Öztürk
  et~al\mbox{.}}{2018}]%
        {_zt_rk_2018}
\bibfield{author}{\bibinfo{person}{Hakime Öztürk}, \bibinfo{person}{Arzucan
  Özgür}, {and} \bibinfo{person}{Elif Ozkirimli}.}
  \bibinfo{year}{2018}\natexlab{}.
\newblock \showarticletitle{DeepDTA: deep drug–target binding affinity
  prediction}.
\newblock \bibinfo{journal}{\emph{Bioinformatics}} \bibinfo{volume}{34},
  \bibinfo{number}{17} (\bibinfo{year}{2018}), \bibinfo{pages}{i821–i829}.
\newblock


\end{thebibliography}

\end{document}